\documentclass[10pt,twocolumn,letterpaper]{article}

\usepackage[pagenumbers]{cvpr} 

\usepackage{algorithm}
\usepackage{algorithmic}
\usepackage{amsmath}
\usepackage{amssymb}
\usepackage{booktabs}
\usepackage{multirow}
\usepackage{graphicx}
\usepackage{tabularx}
\usepackage{array}
\usepackage{cuted}

\renewcommand{\paragraph}[1]{\vspace{.5em}\noindent\textbf{#1.}}

\definecolor{cvprblue}{rgb}{0.21,0.49,0.74}
\usepackage[pagebackref,breaklinks,colorlinks,allcolors=cvprblue]{hyperref}

\def\paperID{} 
\def\confName{CVPR}
\def\confYear{2026}

\title{ColorFLUX: A Structure-Color Decoupling Framework \\ for Old Photo Colorization}

\author{Bingchen Li\textsuperscript{\rm 1}\textsuperscript{*}\textsuperscript{\S} Zhixin Wang\textsuperscript{\rm 2}\textsuperscript{*}\textsuperscript{\dag}
Fan Li\textsuperscript{\rm 2}\textsuperscript{\ddag}
Jiaqi Xu\textsuperscript{\rm 2}
Jiaming Guo\textsuperscript{\rm 2}
Renjing Pei\textsuperscript{\rm 2}
Xin Li\textsuperscript{\rm 1}
Zhibo Chen\textsuperscript{\rm 1}\textsuperscript{\dag}\\
\textsuperscript{\rm 1} University of Science and Technology of China\quad\textsuperscript{\rm 2} Huawei Noah's Ark Lab \\
{\tt\small lbc31415926@mail.ustc.edu.cn, \{xin.li, chenzhibo\}@ustc.edu.cn}\\
{\tt\small \{wangzhixin6, lifan61\}@huawei.com}
}

\begin{document}
\maketitle

\renewcommand\thefootnote{*}
\footnotetext{Equal contribution. \textsuperscript{\dag}Corresponding authors. \textsuperscript{\ddag}Project lead.}
\renewcommand\thefootnote{\S}
\footnotetext{Work done during an internship at Huawei Noah’s Ark Lab.}
\renewcommand\thefootnote{\arabic{footnote}}

\begin{strip}

\centering
\vspace{-65pt}
\includegraphics[width=\linewidth]{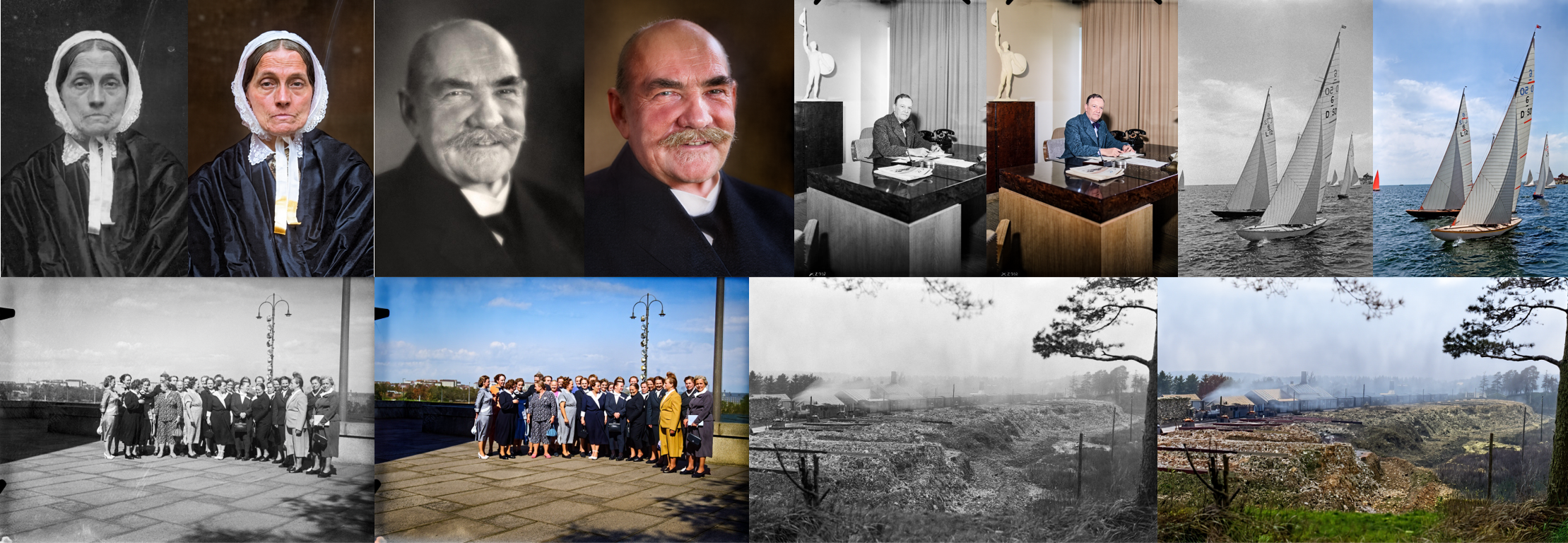}
\vspace{-5mm}

\captionof{figure}{
ColorFLUX achieves vivid and realistic colorization of old photos by decoupling structure and color restoration, exhibiting strong generalization across diverse scenarios, including portraits, group photos, and landscapes.
}
\label{fig:teaser}
\end{strip}

\begin{abstract}

Old photos preserve invaluable historical memories, making their restoration and colorization highly desirable. While existing restoration models can address some degradation issues like denoising and scratch removal, they often struggle with accurate colorization.This limitation arises from the unique degradation inherent in old photos, such as faded brightness and altered color hues, which are different from modern photo distributions, creating a substantial domain gap during colorization.
In this paper, we propose a novel old photo colorization framework based on the generative diffusion model FLUX. Our approach introduces a structure-color decoupling strategy that separates structure preservation from color restoration, enabling accurate colorization of old photos while maintaining structural consistency. We further enhance the model with a progressive Direct Preference Optimization (Pro-DPO) strategy, which allows the model to learn subtle color preferences through coarse-to-fine transitions in color augmentation. Additionally, we address the limitations of text-based prompts by introducing visual semantic prompts, which extract fine-grained semantic information directly from old photos, helping to eliminate the color bias inherent in old photos.
Experimental results on both synthetic and real datasets demonstrate that our approach outperforms existing state-of-the-art colorization methods, including closed-source commercial models, producing high-quality and vivid colorization.
    
\end{abstract}    
\section{Introduction}

\begin{figure}
    \centering
    \includegraphics[width=\linewidth]{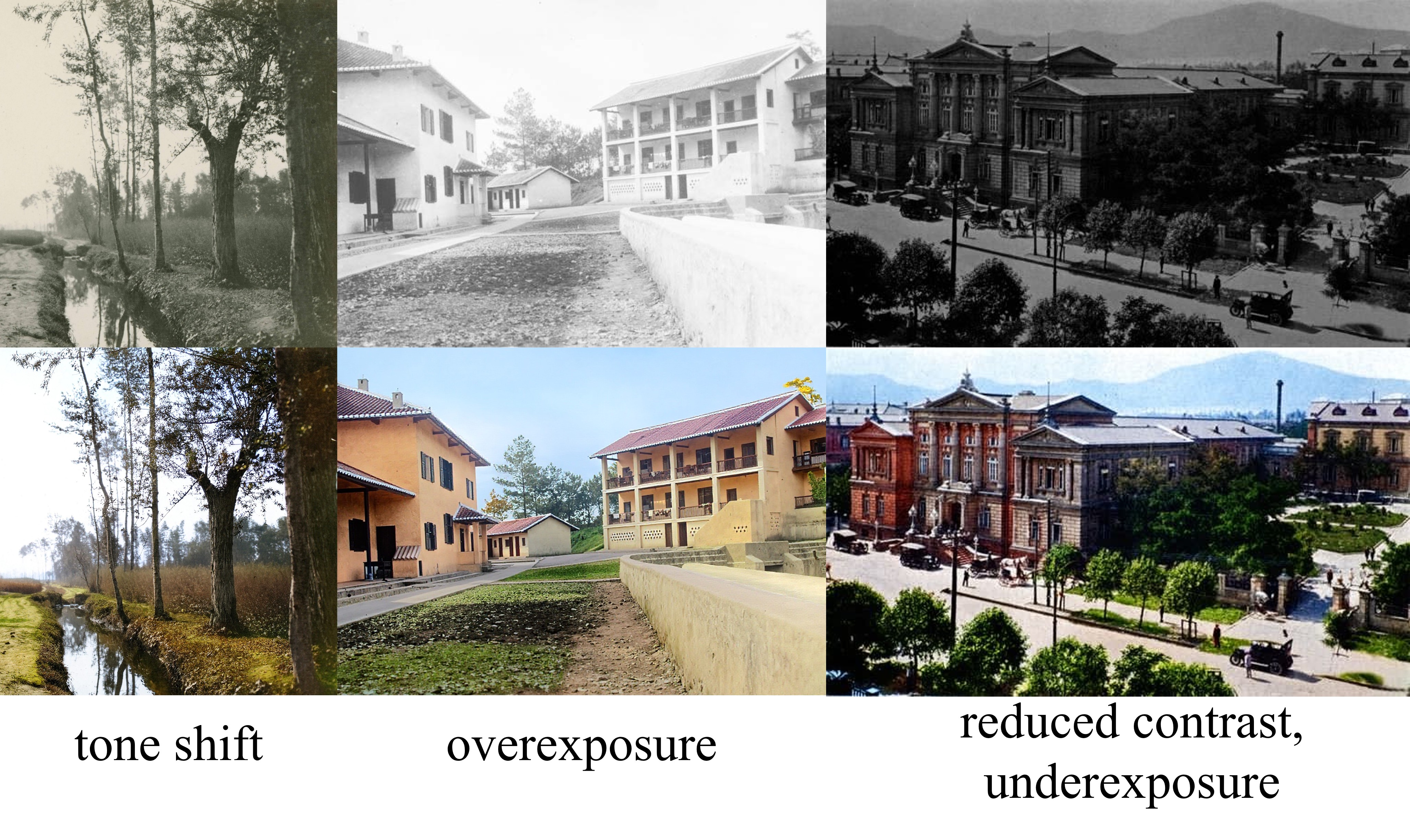}
    \caption{llustration of real old photo’s fading effect. ColorFLUX brings them back to life (second row).}
    \label{fig:realold}
    \vspace{-5mm}
\end{figure}

Old photos serve as invaluable visual records of history, capturing people, places, and moments that define the past. However, due to the limitations of early photographic equipment and preservation conditions, most of these images lack vivid colors (\ie, are black‑and‑white) and often suffer from natural fading (We present some examples in Fig.~\ref{fig:realold}). Recent advances in image restoration models~\cite{yue2023resshift,yu2024scaling,kong2025dpir,qin2025camedit,lin2025jarvisir} have demonstrated strong capabilities in removing common degradations in old photos, such as scratches, noise, and blur. However, these models are not specifically designed for colorization. A practical workaround is to first restore degradations and then apply a designated colorization model~\cite{xu2023pikfix}. But this pipeline faces challenges, as faded grayscale images provide limited color information, requiring the model to infer plausible colors. This involves ensuring semantic consistency, preventing color bleeding, and aligning with modern color preferences. Additionally, the degradation in old photos, like faded brightness and altered color hues, differs from modern photos, creating a difficult training gap. 

Existing generative colorization methods~\cite{liang2024controlcolor} typically leverage text-to-image (T2I) models as color priors while maintaining structural consistency between the grayscale inputs and colorized output, but still struggle with issues like semantic consistency and preventing color bleeding. Moreover, these methods heavily rely on text prompts, which often fail to capture the fine details in image - a limitation encapsulated by the adage, “A picture is worth a thousand words.”

To address these issues, we propose a framework for old photo colorization based on the generative diffusion model FLUX~\cite{flux2024}, chosen for its strong ability to generate high-quality colorized images, even from limited or faded inputs, making it ideal for restoring old photos. The proposed framework introduces an effective structure-color decoupling learning strategy to ensure accurate semantic understanding for color restoration of grayscale old photos. Specifically, the framework consists of three training stages: structure consistency training, basic color learning, and fine-color adjustment. In the structure consistency training stage, FLUX is fine-tuned using ControlNet~\cite{zhang2023addingcontrolnet}, which accepts grayscale images and color visual prompts generated from the color ground truth using Redux\footnote{https://huggingface.co/black-forest-labs/FLUX.1-Redux-dev}. The training loss used is the flow matching loss, enforcing the model to learn the structure rather than the color information.
During the basic color learning stage, we fine-tune the pretrained Redux model, allowing it to extract natural semantic embeddings that connect the priors of FLUX, while abandoning the color bias from old photo inputs. This stage combines flow matching loss and feature distillation loss, while freezing both the structure-preserving ControlNet and the pretrained FLUX, to decouple it from the structural training.

After restoring the basic colors, the next step is to correct fading effects such as low saturation or over-/under-exposure, which are commonly observed in old photos. We propose fine-color adjustment, a post-training process implemented via Direct Preference Optimization (DPO), enhancing the model's ability to generate vivid colors that align with modern aesthetics and human preferences. Specifically, the preference dataset for DPO training consists of triplets: input, positive, and negative samples. Positive samples are high-quality, realistic images, while negative samples are their corresponding color-augmented versions, and the input is the grayscale version of the color-augmented images. However, learning directly from a large color-augmentation space can overwhelm the model, causing it to focus on obvious color changes rather than more subtle adjustments. To address this, we propose a progressive DPO (Pro-DPO) strategy that begins with larger, more noticeable augmentations and gradually transitions to finer ones, helping the model learn more subtle color preferences.

Experimental results for both real and synthetic datasets demonstrate that our approach outperforms both existing state-of-the-art colorization models and closed-source commercial solutions.
We summarize our contributions as:
\begin{itemize}
    \item We present a framework for old photo colorization based on FLUX and incorporates an effective structure-color decoupling learning strategy.
    \item We propose a progressive DPO strategy that post-trains the model through coarse-to-fine color augmentation transitions, enabling the model to learn subtle color preferences and generate vivid colors for old photos.
    \item We conduct extensive experiments to evaluate our colorization framework, demonstrating that it achieves state-of-the-art performance, outperforming both existing methods and closed-source commercial solutions.
\end{itemize}

\section{Related Works}

\begin{figure*}[t]
    \centering
    \includegraphics[width=1\linewidth]{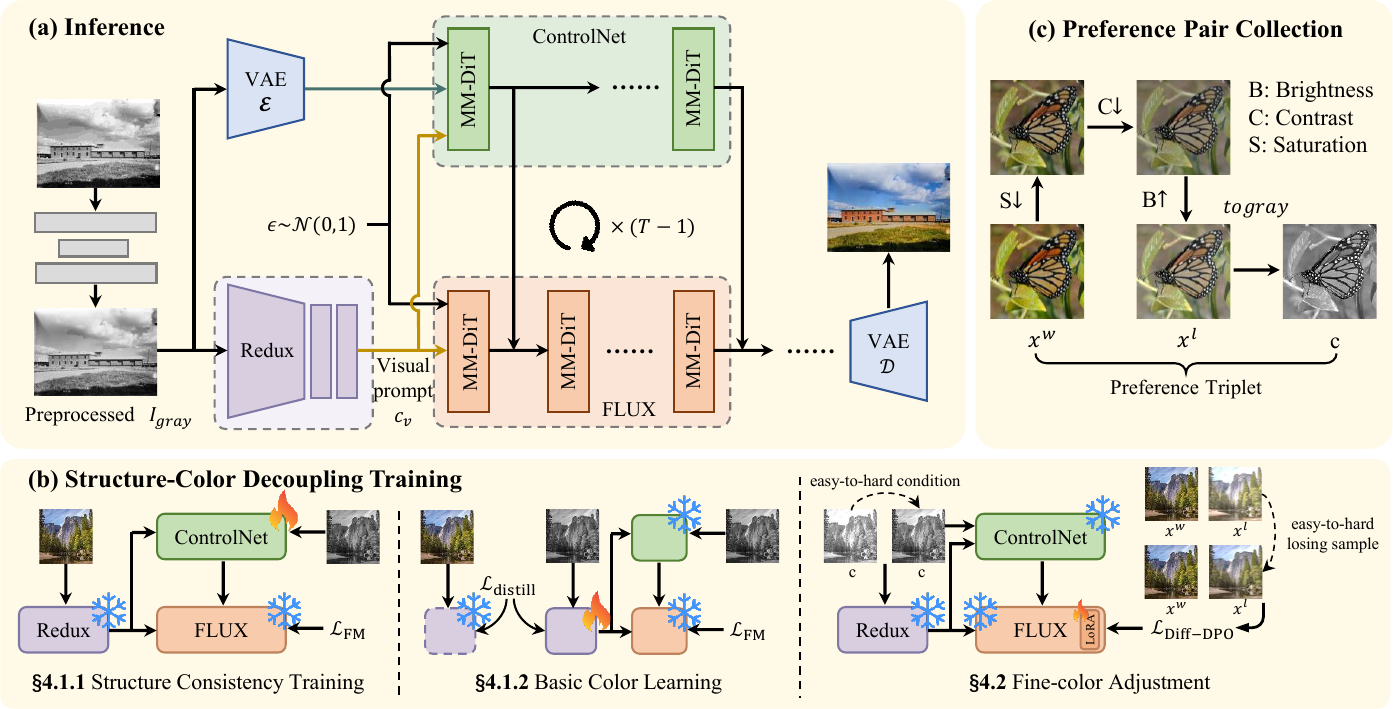}
    \caption{The framework of our proposed method. (a) The inference process of colorization, including a preprocessor to remove degradations in old photos. (b) The training procedure of our method. Notably, we illustrate an example of easy-to-hard losing samples. The model progressively learns preferences from severely augmented samples to milder ones, ensuring precise color perception to effectively counteract the fading effects of old photos. (c) The preference pair collection pipeline. To synthesize $x^l$, we randomly apply augmentation combinations sampled from [`B', `C', `S'].}
    \vspace{-4mm}
    \label{fig:framework}
\end{figure*}

\subsection{Old Photo Restoration}

The pioneer work~\cite{wan2020bringing} performs transfer learning in the latent space to align the distribution of old photos with that of clean images, effectively removing degradations present in old photos. However, it falls short in colorization, as it lacks explicit mechanisms to model vivid colors from the limited information available in grayscale inputs. DeOldify~\cite{deoldify} leverages GANs to learn the colorization of old photos, which utilizes self-attention in the U-Net generator to extract color cues from grayscale images. Pik-Fix~\cite{xu2023pikfix} adopts a divide-and-conquer strategy, restoring old photos through restoration, similarity, and colorization sub-networks, with a reference image providing chromatic guidance. 

\subsection{Image Colorization}
Image colorization typically aims to add color to a grayscale image, restoring the color information that has either been lost or was originally absent. Early works~\cite{deshpande2015learning,iizuka2016let,cong2024automatic} primarily focused on automatic colorization, where networks learn pixel-wise colorization mappings from synthetic color–grayscale image pairs. However, due to the limited chromatic cues available in grayscale images, these automatic methods often struggle to generate rich and realistic colors. Furthermore, they face challenges in accurately understanding the semantics of image instances, resulting in artifacts such as color bleeding~\cite{liang2024controlcolor}. To mitigate these issues, a series of works~\cite{wang2022palgan,weng2023Lcad,liang2024controlcolor,jarvisart2025,lin2025jarvisevo} propose to leverage generative priors to better understand image semantics, while employing various forms of guidance to direct colorization process, such as color hints~\cite{yun2023icolorit,cho2023guiding}, exemplar images~\cite{bozic2024versatileexemplar,bugeau2013variational,carrillo2022super}, or language descriptions~\cite{weng2023Lcad,chang2023lcoins,weng2022code,chang2022coder,weng2022ct}, \etc

However, directly applying these colorization methods to old photos faces a major challenge: the fading effects commonly present in old photos can lead to colorization results that do not align with human preferences, particularly in color attributes such as brightness, contrast, and saturation, which are highly perceptible to the human eye. In this paper, we address this issue by introducing a color-aware progressive Direct Preference Optimization (DPO) learning strategy within the colorization framework.

\subsection{Direct Preference Optimization for Diffusion}

Direct Preference optimization (DPO)~\cite{rafailov2023directdpo} can be applied to pre-trained diffusion models to align them with human preferences. Diffusion-DPO~\cite{wallace2024diffusiondpo} firstly introduces DPO into T2I generation and fine-tunes diffusion models with preference data collected from human-annotated Pick-a-Pic~\cite{pickapic} dataset, achieving better prompt alignment and aesthetic scores. Diffusion-RPO~\cite{gu2024diffusionrpo} improves the construction of the preference datasets by introducing relative samples with similar semantics. MaPO~\cite{mapo} handles reference mismatch by replacing divergence regularization on the reference model with an amplification factor. D3PO~\cite{yang2024usingd3po} further adapts DPO to discrete diffusion models for structured sequence generation tasks, ensuring preference alignment without explicit reward models.
\section{Preliminaries}

\subsection{Rectified Flow-based Diffusion}

Modern generative modeling~\cite{esser2024scalingsd3,flux2024} are usually based on Rectified Flow~\cite{liu2022rectifiedflow}, which defines the intermediate noisy data $x_t$ by:
\begin{equation}
    x_t = (1-t)x_0 + t \epsilon,
\end{equation}
where $t \in [0,1]$ is the sampled timestep, $x_0 \sim p(x_0)$ is drawn from real data distribution $p(x_0)$, and $\epsilon\sim\mathcal{N}(0,I)$ is standard Gaussian noise. The predicted velocity field $v_\theta(x_t,t)$ by the diffusion transformer can be optimized by the flow matching objective:
\begin{equation}
    \label{eq:velocity}
    \mathcal{L}_{\text{FM}} = \mathbb{E}_{t,\, x_0 \sim p(x_0),\, \epsilon \sim \mathcal{N}(0,I)} \left[ \left\| v - v_\theta(x_t, t) \right\|^2 \right],
\end{equation}
where the target velocity field is $v=\epsilon-x_0$.

\subsection{Diffusion Direct Preference Optimization}

Preference optimization aims to train models using triplets $(c,x^w,x^l) \sim D$, where the model is encouraged to generate preferred outputs $x^w$ over dispreferred ones $x^l$ under a given condition $c$ (\eg, text prompt). Rafailov \etal~\cite{rafailov2023directdpo} propose Direct Preference Optimization (DPO), which bypasses explicit reward modeling and instead directly optimizes the model parameters based on preference pairs. To adapt DPO to diffusion models, Wallace \etal~\cite{wallace2024diffusiondpo} replace policy log‑likelihoods with diffusion ELBO scores and derive $\mathcal{L}_{\text{Diff-DPO}}$ as:
\begin{multline}
- \mathbb{E} \Bigg[ \log \sigma \Bigg( 
  -\frac{\beta}{2} \Big(
    \left\| \epsilon^w - \epsilon_{\theta}(x_t^w, t) \right\|^2 
    - \left\| \epsilon^w - \epsilon_{\text{ref}}(x_t^w, t) \right\|^2 \\
    {} - \left(
      \left\| \epsilon^l - \epsilon_{\theta}(x_t^l, t) \right\|^2 
      - \left\| \epsilon^l - \epsilon_{\text{ref}}(x_t^l, t) \right\|^2
    \right)
  \Big)
\Bigg) \Bigg],
\end{multline}
where the expectation is taken over preference triplets and the noise schedule $t$. Consequently, to extend $\mathcal{L}_{\text{Diff-DPO}}$ to advanced Rectified Flow-based diffusion models, Liu \etal~\cite{liu2025improvingflowdpo,liu2025flowgrpo} replace the noise vector $\epsilon^*$ with the velocity field $v^*$ in Eq.~\ref{eq:velocity} and form the final loss function (denote as $\mathcal{L}_{\text{Diff-DPO}}$ in the following paper):
\begin{multline}
- \mathbb{E} \Bigg[ \log \sigma \Bigg( 
  -\frac{\beta_t}{2} \Big(
    \left\| v^w - v_\theta(x_t^w, t) \right\|^2
    - \left\| v^w - v_{\text{ref}}(x_t^w, t) \right\|^2 \\
    {} - \left(
      \left\| v^l - v_\theta(x_t^l, t) \right\|^2
      - \left\| v^l - v_{\text{ref}}(x_t^l, t) \right\|^2
    \right)
  \Big)
\Bigg) \Bigg],
\end{multline}
where $\beta_t=\beta(1-t)^2$. Moreover, Liu \etal~\cite{liu2025flowgrpo} demonstrate that using a constant $\beta_t=\beta_c$ achieves better results.
\section{Method}

In this paper, we propose a novel colorization framework for old photos based on FLUX\cite{flux2024}, which consists of three training stages: structure consistency training, basic color learning, and fine-color adjustment. Moreover, as shown in Figure~\ref{fig:framework}, we utilize a pretrained low-level image preprocessor to remove degradations such as noise, and blur, thus eliminating potential interference with the subsequent colorization process. For simplicity, we denote the preprocessed image as $I_{gray}$ in the remainder of this paper. Additionally, to comprehensively evaluate our method, we collect 50 real old photos from the Internet and develop an evaluation metric based on a multimodal large language model, as detailed in Sec.~\ref{sec:evalution}.

\subsection{Decoupling Structure-Color Learning}
\label{sec:structure}
\subsubsection{Structure Consistency Training} 

Following the common practice~\cite{liang2024controlcolor}, we utilize ControlNet~\cite{zhang2023addingcontrolnet} to inject structural information into FLUX. Unlike previous methods, our ControlNet takes two inputs simultaneously: the grayscale image and a color visual prompt, as shown in Fig.~\ref{fig:framework} (b). The latter input is obtained by extracting color embeddings from $I_{gt}$ using a pretrained Redux model, providing the model with accurate color priors. This training approach encourages the model to primarily focus on structural consistency rather than color restoration at this stage. Furthermore, we use the flow matching loss defined in Eq.~\ref{eq:velocity} for training, while keeping the pretrained FLUX and Redux models frozen during this stage.

\subsubsection{Basic Color Learning} 

In this stage, we fully leverage the color prior knowledge embedded in the pretrained FLUX model to infer plausible basic colors, aligning them with modern human preferences while preventing color bleeding. Existing approaches relying on textual prompts may fall short in capturing fine-grained details, making them inadequate for preventing color bleeding in old photo colorization tasks. To address this, as illustrated in Fig.~\ref{fig:framework}(b), we fine-tune the pretrained Redux model to extract natural semantic embeddings closely aligned with the color priors of FLUX, while simultaneously reducing the influence of color biases inherent in old photos. Specifically, we fine-tune the Redux model $\Phi$ through color-knowledge distillation from the $I_{gt}$ to the grayscale input image $I_{gray}$ as follows:
\begin{equation}
    \mathcal{L}_{\text{distill}}=\|\Phi(I_{gray})-\Phi'(I_{gt})\|^2,
\end{equation}
where $\Phi'$ denotes a copy of pretrained Redux model that is frozen and specifically used to extract ground-truth color information. This distillation loss drives the embeddings extracted from grayscale images to be close to the image embedding extracted from their corresponding color images.
To ensure that the generated visual semantic prompt is well aligned with the color priors space of the pre-trained FLUX model, we combine both $\mathcal{L}_{\text{FM}}$ and $\mathcal{L}_{\text{distill}}$ to fine-tune $\Phi$.

\begin{equation}
    \mathcal{L} = \mathcal{L}_{\text{FM}} + \alpha\mathcal{L}_{\text{distill}}
\end{equation}
where $\alpha$ is a weighting hyperparameter.

\begin{table*}[t]
\centering
\caption{Quantitative comparisons across different methods on two benchmarks. VQ-R1, CRI, CRA, CCS, SCS, AES, and OA denote VisualQuality-R1, Color RIchness, Color RAtionality, Color ConSistency, Structure ConSistency, AESthetics, and OverAll performance, respectively. Notably, we apply our preprocessor to all methods to ensure a fair comparisons on colorization performance.  
The Runtime is measured on a single GPU with about 19.5 TFLOPS.
Best results are \textbf{bolded.}}
\label{tab:maintable}
\setlength{\tabcolsep}{9pt}
\resizebox{1\textwidth}{!}{
\begin{tabular}{lcccccccccc}
\toprule
             & \multicolumn{3}{c}{NR-IQA} & \multicolumn{6}{c}{Qwen-score} & \multirow{2}{*}{Runtime}       \\ \cmidrule(l){2-4} \cmidrule(l){5-10} 
             &    DeQA                           &    Q-Insight                                 &                             VQ-R1      & CRI   & CRA   & CCS   & SCS   & AES & OA   \\ \midrule
\multicolumn{11}{c}{\textit{DIV2K-valid-synthesized}}                                                                                                                                            \\ \midrule

DeOldify~\cite{deoldify} & 4.080 & 3.978 & 4.625 & 69.35 & \textbf{80.95} & 83.80 & \textbf{94.78} & 80.70 & 81.73 & 2.23s \\

DDColor~\cite{kang2023ddcolor} & 4.162 & 3.992 & 4.574 & 79.80 &	78.55 & 77.60 & 93.22 & 81.13 & 81.80 & 0.19s \\

CtrlColor~\cite{liang2024controlcolor} & 4.144 & \textbf{4.013} & 4.654 & 75.40 & 80.25 & \textbf{84.04} & 94.53 & 81.85 & 83.28 & 8.05s\\

FLUX-Kontext~\cite{labs2025flux1kontextflowmatching} & 4.209 & 4.002 & 4.661 & 77.75 & 80.00 & 83.61 & 93.91 & 82.22 & 83.48 & 28.14s\\

ColorFLUX (Ours) & \textbf{4.293} & 4.008 & \textbf{4.739} & \textbf{80.80} & 79.70 & 82.23 & 94.05 & \textbf{83.12} & \textbf{83.73} & 7.45s\\
\midrule

\multicolumn{11}{c}{\textit{DIV2K-valid-augmented}}                                                                                                                                            \\ \midrule
DeOldify~\cite{deoldify}                    & 3.841                 & 3.798                       & 4.349                      & 64.70 & 77.75 & 81.37 & 94.29 &  77.63 & 79.11 & 2.23s\\
DDColor~\cite{kang2023ddcolor}               &            3.941           &    3.877                        &    4.410                               &    77.85   &  76.70     &   78.66    &   93.43    &      80.43 & 81.54  & 0.19s \\
CtrlColor~\cite{liang2024controlcolor}          &           3.917            &       3.857                     &        4.432                           &     72.35  &   78.40    &   82.99    &    94.03   &    80.25  & 81.71 & 8.05s \\
FLUX-Kontext~\cite{labs2025flux1kontextflowmatching} & 4.175 & \textbf{3.981} & 4.617 & 77.25 & \textbf{79.90} & \textbf{84.57} & 94.63 & 81.83 & 82.55 & 28.14s \\

ColorFLUX (Ours)             &        \textbf{4.303}               &                 3.978           &        \textbf{4.728}                           &  \textbf{80.15}     &    79.70   &    82.66   &   \textbf{94.83}    &     \textbf{82.80} & \textbf{83.53} & 7.45s
\\ \midrule
\multicolumn{11}{c}{\textit{RealOldPhotos}}                                                                                                                                                  \\ \midrule
DeOldify~\cite{deoldify}                & 4.056                  & 3.852                      & 4.419                            & 70.90 & 80.50 & 82.66 & 94.76 & 79.02 & 80.86 & 2.23s \\
DDColor~\cite{kang2023ddcolor}               &            4.090           &      \textbf{3.935}                      &      4.500                             &   76.20    &    81.10   &   82.36    &   94.60    &    82.42 & 82.70  & 0.19s  \\
CtrlColor~\cite{liang2024controlcolor}            &           4.050            &    3.931                        &        4.486                           &    76.40   &   80.50    &    83.24   &   94.50    &   81.64 & 82.68  & 8.05s \\
FLUX-Kontext~\cite{labs2025flux1kontextflowmatching} &          4.110           &       3.843                     &        4.452                           &   66.00    &   76.90    &     79.74  &  \textbf{94.88}    &   76.38 & 77.70 & 28.14s \\
Doubao~\cite{doubao}           &          3.741             &      3.563                      &             4.097                      &    \textbf{83.10}   &   64.90    &   75.10    &   89.00    &     74.40 & 75.24   & $\sim$29s\footnotemark \\
ColorFLUX (Ours)               &    \textbf{4.199} &  3.869 &  \textbf{4.593} &  80.50           &     \textbf{81.30}                       &      \textbf{83.36}                             &    94.46   &   \textbf{83.22} & \textbf{83.20}  & 7.45s   \\ \bottomrule
\end{tabular}%
}
\vspace{-4mm}
\end{table*}

\subsection{Progressive DPO for Fine-color Adjustment}
\label{sec:DPO}
While basic color restoration provides a foundation, it is insufficient to fully revive old photos. The next step is to subtly correct common fading effects in old photos, such as low saturation, over-/under-exposure. 
Direct Preference Optimization (DPO) is a widely used post-training method that leverages and enhances a model’s inherent ability to generate vivid colors aligned with modern aesthetics and human preferences. In this paper, we construct a color-preference dataset and introduce a progressive DPO training strategy for fine-color adjustment.

\subsubsection{Color-Preference Pairs Collection} 
The color-preference dataset for DPO training consists of triplets: input, positive, and negative samples. Intuitively, realistic and naturally colored images are defined as the winning (positive) samples denoted as $x^w$ while losing (negative) samples, denoted as $x^l$ are synthesized from $x^w$ by simulating various fading artifacts. Formally, given a colored ground-truth image $x^w=I_{gt}$, we generate the losing sample  $x^l=I_{aug}$ with randomly shuffled combinations of \textit{brightness adjustment}, \textit{contrast reduction}, and \textit{saturation reduction}. The condition $c$ is obtained by applying the \texttt{RGB2GRAY} algorithm to $I_{aug}$. The overall pipeline is illustrated in Fig.~\ref{fig:framework} (c).

\subsubsection{Progressive DPO Training} 
Learning directly from a wide range of color augmentations may cause the model to overfit to simple preference patterns (e.g., avoiding excessively saturated results) while neglecting more subtle but valuable fine-color adjustments, such as slightly increasing saturation to enhance vividness. To mitigate this, we propose a progressive DPO (Pro-DPO) strategy that gradually shifts the training focus from coarse to fine adjustments, enabling the model to capture subtle color preferences over time.
Specifically, we design a two-stage progressive DPO training scheme. In the first stage, we apply strong augmentations to construct obvious preference gaps, allowing the model to learn coarse distinctions. In the second stage, we gradually reduce the augmentation strength, introducing more subtle distinctions and encouraging the model to refine its ability to perform fine-grained color adjustments.

\footnotetext{Runtime is measured through its official website interface.}

\begin{figure*}
    \centering
    \includegraphics[width=1\linewidth]{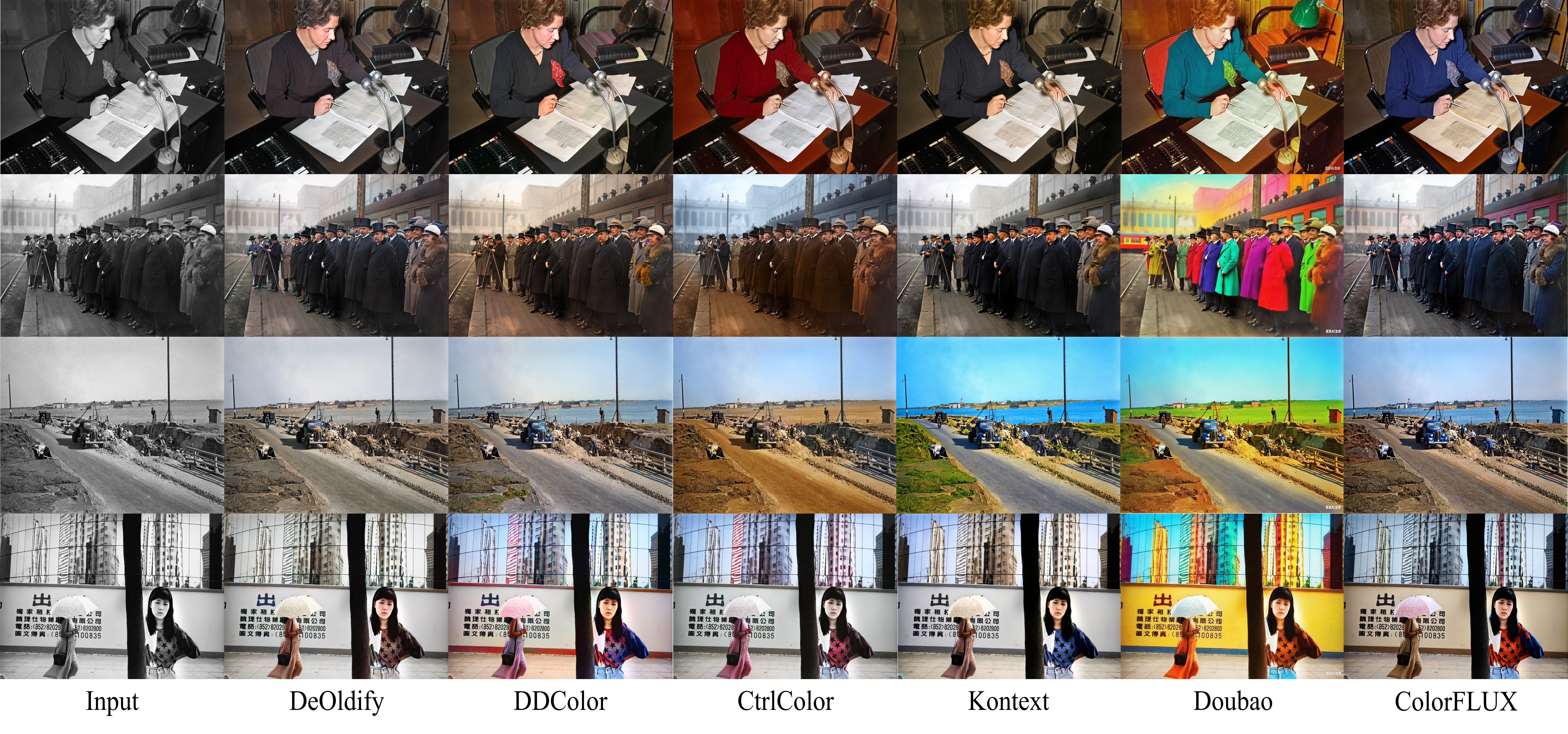}
    \vspace{-5mm}
    \caption{Qualitative comparisons between ColorFLUX and other methods on \textit{RealOldPhotos}. Zoom in for better views.}
    \vspace{-4.5mm}
    \label{fig:allresults}
\end{figure*}

\subsection{Evaluation of Colorization Framework}
\label{sec:evalution}

Assessing the quality of colorized images is particularly challenging, as color is inherently more subjective and ambiguous compared to image content. Although numerous Image Quality Assessment (IQA) metrics exist~\cite{yang2022maniqa,qalign}, there is still no reliable metric specifically tailored to evaluate image colorization. Inspired by evaluation protocols in multimodal large language models (\eg, GPT-Score)~\cite{tong2025gptscore}, we employ the open-source Qwen2.5-VL-72B model~\cite{Qwen2.5-VL} to quantitatively assess the perceptual quality of colorized results. The detailed prompts are provided in Sec.~\ref{sec:qwen}. 
Moreover, prior works~\cite{wan2020bringing,xu2023pikfix} primarily focus on restoring portrait-style old photos with limited samples (\eg, only 13 old photos in DDColor~\cite{kang2023ddcolor}), while largely overlooking more complex scenarios such as crowds or natural landscapes. To enable a more comprehensive evaluation of colorization performance, we additionally collected a set of 50 publicly available old photos from the Internet~\footnote{\url{https://huggingface.co/datasets/NatLibFi/Finna-JOKA-images}} (\ie, \textit{RealOldPhotos}), covering diverse contents.
\section{Experiments}

\subsection{Implementation Details}
\paragraph{Network Architecture} We build our colorization framework based on FLUX.1~\cite{flux2024}. We adopt the official FLUX Redux model~\cite{flux2024} as the initialization of our semantic visual prompt extractor, which consists of a SigLip~\cite{zhai2023sigmoidsiglip} vision encoder and a two-layer MLP to align the features to the text branch of FLUX. For DPO training, we apply LoRA with a rank of 32 to the attention and feed-forward layers within all MM-DiT blocks of FLUX. For preprocessor, we adopt the pre-trained S3Diff~\cite{zhang2024degradations3diff}.

\paragraph{Training Details} The training is conducted on $8\times 80$ GB GPUs and completes in around two days. We provide detailed training descriptions in Sec.~\ref{sec:details}.

\paragraph{Inference} During inference, we resize the input to $1024\times 1024$ (same as training), use the Euler flow-matching noise scheduler, and fix the random seed, guidance scale, and control scale to 0, 3.5, and 1, respectively, with a total of 8 sampling steps. After denoising, we resize the image back to the original resolution.
 
\paragraph{Metrics} As mentioned in Sec.~\ref{sec:evalution}, we primarily use Qwen2.5-VL-72B\footnote{https://huggingface.co/Qwen/Qwen2.5-VL-72B-Instruct} (abbreviated as Qwen-score) to evaluate the performance of colorization. We focus on six aspects: Color RIchness (CRI), Color RAtionality (CRA), Color ConSistency (CCS), Structural ConSistency (SCS), AESthetics (AES), and OverAll (OA). As a supplement to Qwen-score, we further adopt three state-of-the-art NR-IQA metrics, including DeQA~\cite{you2025teachingdeqa}, Q-Insight~\cite{li2025qinsight}, and VisualQuality-R1~\cite{wu2025visualquality}.

\begin{figure}[!h]
    \centering
    \includegraphics[width=1\linewidth]{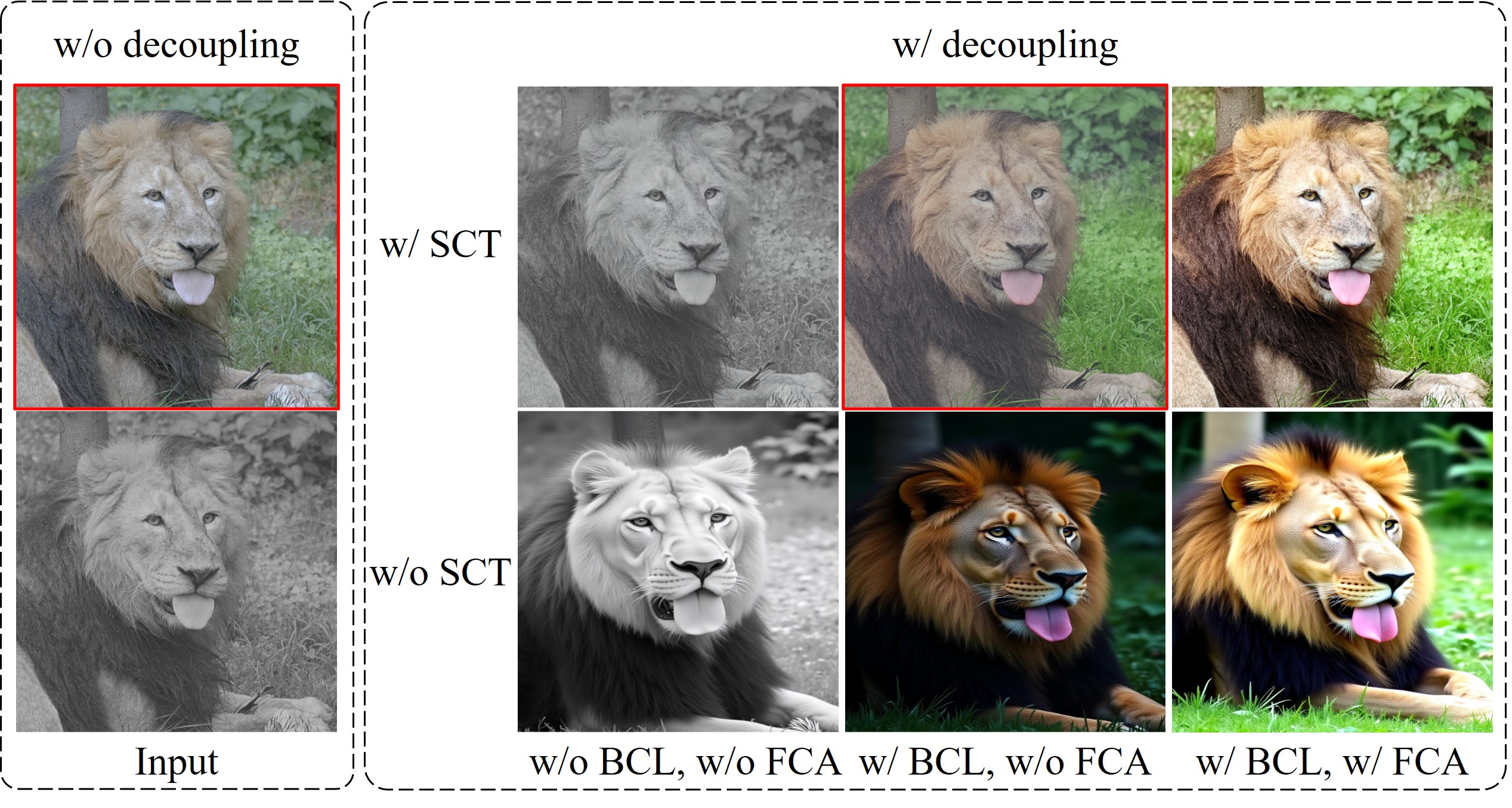}
    \caption{Visual results of decoupling each parts of ColorFLUX. We denote structure consistency training, basic color learning, and fine-color adjustment by SCT, BCL, and FCA, respectively. ``W/o'' represents that we remove the according network from the trained framework (\eg, w/o SCT denotes for removing ControlNet).}
    \vspace{-4mm}
    \label{fig:decoupling}
\end{figure}

\paragraph{Evaluation Datasets} We evaluate the performance of our method on both synthesized and real datasets. For the former, we adopt two settings: (i) we convert the DIV2K validation set (100 images) to grayscale using the \texttt{RGB2GRAY} algorithm (\textit{DIV2K-valid-synthesized}); and (ii) we first apply color augmentations—such as brightness adjustment, saturation reduction, and contrast reduction—to the original images, and then convert them to grayscale (\textit{DIV2K-valid-augmented}). The first setting follows the evaluation protocol adopted by previous image colorization works~\cite{ozbulak2019image,kang2023ddcolor} for fair comparison, whereas the second setting is introduced to mimic the color fading and degradation artifacts typically present in old photos. For the latter, we use the old photos we collected from the Internet  (\textit{RealOldPhotos}).

\subsection{Colorization Results}

For synthesized datasets, we compare our method with three representative colorization approaches: an open-source tool, DeOldify~\cite{deoldify}; an encoder–decoder-based method, DDColor~\cite{kang2023ddcolor}; a diffusion-based method, CtrlColor~\cite{liang2024controlcolor}; and a FLUX-based editing model, FLUX-Kontext~\cite{labs2025flux1kontextflowmatching}. For the real dataset, we additionally include a closed-source image editing model, Doubao, to assess ColorFLUX’s performance in comparison with a commercial-grade model.

We demonstrate the quantitative results in Tab.~\ref{tab:maintable}. As observed, ColorFLUX achieves the best performance on DeQA, VisualQuality-R1 (VQ-R1), Aesthetics (AES), and the overall (OA) score across all three benchmarks. Compared with DDColor and CtrlColor, ColorFLUX not only demonstrates higher color richness but also achieves superior color rationality. This suggests that our method is capable of enriching old photos with vivid colors, while effectively leveraging the pre-trained generative prior through our visual semantic prompts to produce colors that align more naturally with real-world scenes. Notably, although Doubao achieves higher color richness than ColorFLUX, its results tend to be overly saturated, leading to unrealistic appearances as depicted in Fig.~\ref{fig:allresults}. This further demonstrates that our method not only enhances color richness but also preserves color rationality.

Referring to Fig.~\ref{fig:allresults}, ColorFLUX consistently produces visually appropriate colorization results against different old photos. In contrast, other methods exhibit large variations in brightness (\eg, Kontext), or produce overly dull and monochromatic results due to the fading effect (\eg, DeOldify and CtrlColor). Thanks to the accurate semantics extracted by our visual semantic prompt, ColorFLUX also maintains strong color consistency in complex scenes. In comparison, DDColor and DeOldify suffer from noticeable color bleeding artifacts (\eg, hands and coats).

\begin{table}[t]
\centering
\caption{Quantitative comparisons across different semantic prompt strategies. ``VS Prompt'' refers to visual semantic prompt. Best results are \textbf{bolded.}}
\label{tab:abprompt}
\resizebox{\linewidth}{!}{%
\begin{tabular}{@{}lccccccc@{}}
\toprule
                   &  DeQA & Q-Insight& VQ-R1 &CRI   & CRA   & CCS   & AES   \\ \midrule
\multicolumn{8}{c}{\textit{DIV2K-valid-augmented}}                         \\ \midrule
Text Prompt    & 3.758 & 3.689 & 4.289 & 50.00 & 68.70 & 78.82 & 69.74 \\
VS Prompt   & \textbf{3.932} & \textbf{3.752} & \textbf{4.344} & \textbf{55.80} & \textbf{71.85} & \textbf{79.44}& \textbf{72.24} \\ \midrule
\multicolumn{8}{c}{\textit{RealOldPhotos}}                              \\ \midrule
Text Prompt    & 3.863 & 3.761 & 4.248 & 58.00 & 71.70 & 75.00 & 70.06 \\
VS Prompt   & \textbf{4.105} & \textbf{3.811} & \textbf{4.381}& \textbf{67.80} & \textbf{79.60} & \textbf{82.16}& \textbf{78.62}   \\ \bottomrule
\end{tabular}%
}
\vspace{-3mm}
\end{table}

\begin{figure}
    \centering
    \includegraphics[width=1\linewidth]{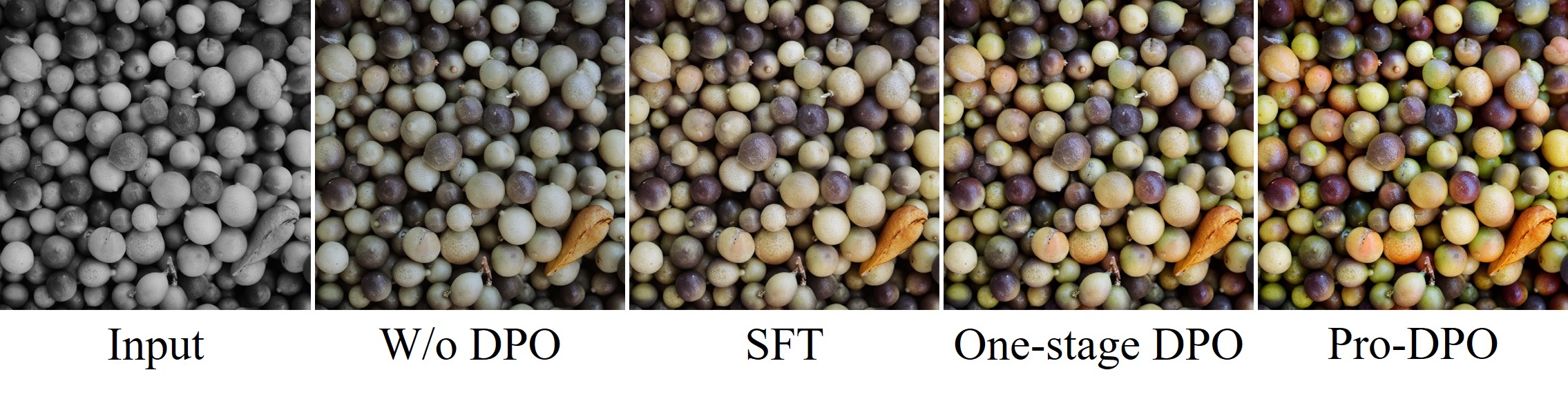}
    \caption{Qualitative comparisons for different fine-color post training strategy on \textit{DIV2K-valid-augmented}.}
    \label{fig:abdpo}
\end{figure}

\subsection{Ablation Studies}

\paragraph{Effectiveness of Structure-Color Decoupling} We investigate the effectiveness of our structure-color decoupling strategy by visualizing the contributions of each network component to the final colorization results. As shown in the right part of Fig.~\ref{fig:decoupling}, removing the SCT-optimized ControlNet leads to severe structural inconsistency. While incorporating the BCL-optimized Redux recovers basic colorization, the results still suffer from color degradations inherited from the grayscale input. With the addition of the FCA-optimized LoRA weights, ColorFLUX finally achieves visually pleasing colorization results with enhanced fidelity and vividness.

In the left part of Fig.~\ref{fig:decoupling}, we show results from jointly training Redux and ControlNet without structure-color decoupling. Comparing the two red-boxed images, the non-decoupled result shows a noticeably duller facial region on the lion, indicating that entangling color learning and structural preservation may interfere with each other, leading to suboptimal results.

\begin{table}[t]
\centering
\caption{Quantitative comparisons across different basic color learning strategies. ``$\mathcal{L}_{\text{FM}}$'' and ``$\mathcal{L}_{\text{distill}}$'' denote flow matching loss and distillation loss, respectively. We report results on \textit{DIV2K-valid-synthesized} without augmentations.}
\label{tab:suppabredux}
\resizebox{\linewidth}{!}{%
\begin{tabular}{@{}lccccccc@{}}
\toprule
        
              &  DeQA & Q-Insight& VQ-R1 &CRI   & CRA   & AES   & OA  \\ \midrule
$\mathcal{L}_{\text{FM}}$       & 3.889 & 3.868 & 4.432 & 46.35 & 68.45 & 67.64& 69.91 \\
$\mathcal{L}_{\text{distill}}$           &  4.044    &   3.895    &  4.552     &    \textbf{58.55}   &   70.65    &   72.04 & 75.12    \\
$\mathcal{L}_{\text{FM}}+\mathcal{L}_{\text{distill}}$ &   \textbf{4.114}    &  \textbf{3.907}   &    \textbf{4.577}   &  57.55    &  \textbf{72.45}     &   \textbf{73.32} & \textbf{75.39}   \\\bottomrule
\end{tabular}%
}
\end{table}

\begin{figure}
    \centering
    \includegraphics[width=1\linewidth]{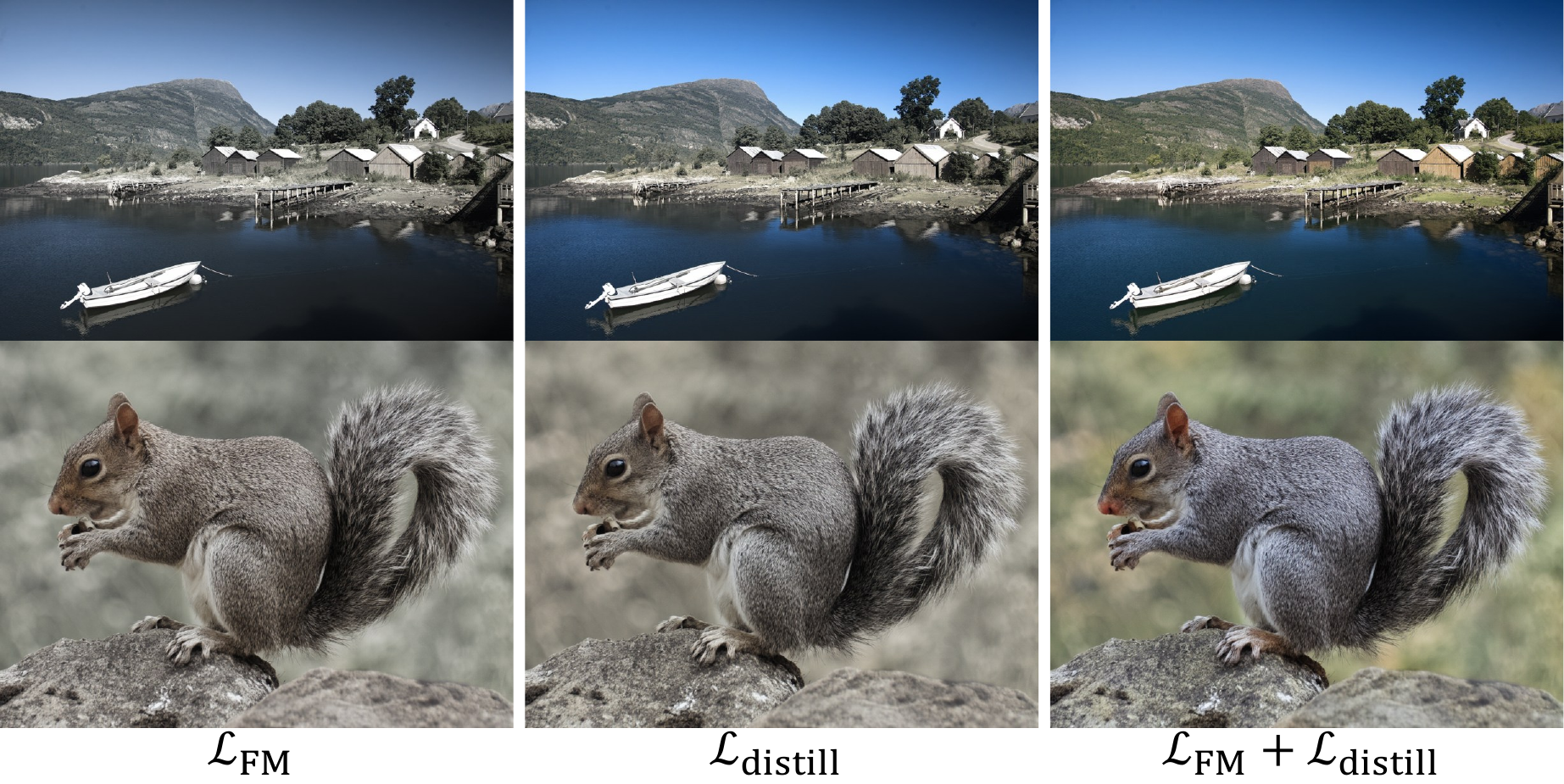}
    \caption{Visual comparisons between different basic color learning strategies on \textit{DIV2K-valid-augmented}.}
    \label{fig:suppredux}
    \vspace{-2mm}
\end{figure}

\paragraph{Effectiveness of $\mathcal{L}_{\text{FM}}$ and $\mathcal{L}_{\text{distill}}$} In this part, we study the effectiveness of utilizing a combination of flow matching loss $\mathcal{L}_{\text{FM}}$ and distillation loss $\mathcal{L}_{\text{distill}}$ to perform the basic color learning of the Redux model. Specifically, we compare our implementation with two more settings: (i) only use $\mathcal{L}_{\text{FM}}$, and (ii) only use $\mathcal{L}_{\text{distill}}$ with out FLUX involving. As demonstrated in Tab.~\ref{tab:suppabredux}, our implementation achieves the best NR-IQA scores and almost the best Qwen-scores. Specifically, using only $\mathcal{L}_{\text{FM}}$ ensures that Redux’s output aligns with the pre-trained representation space of FLUX, but may lead to low training efficiency, resulting in suboptimal distillation performance. On the other hand, although distill the Redux model alone achieves better color richness, the relatively lower color rationality suggests a potential distribution shift in Redux’s output space without the supervision of $\mathcal{L}_{\text{FM}}$, which impairs FLUX’s ability to accurately interpret the visual semantic prompt. This is also validated through visualization in Fig.~\ref{fig:suppredux}, where our implementation achieves the most vibrant colorization results. 

\paragraph{Effectiveness of Visual Semantic Prompt} We compare our visual semantic prompt design (before DPO training) with the commonly used text prompt approach in Tab.~\ref{tab:abprompt}. For text prompt training, we utilize Qwen2.5-VL-72B to extract detailed descriptions from images, while filtering out phrases such as “black-and-white image” to avoid incorrect semantic propagation. As demonstrated, leveraging visual semantic prompts consistently outperforms text prompts across all evaluation metrics on both benchmarks. The improvement is especially significant on \textit{RealOldPhotos}, as text prompts often fail to capture the full semantic context—particularly in complex, degraded scenes such as old photos—leading to reduced color richness (CRI) and inaccurate color assignments (low CRA) due to insufficient semantic guidance.

\begin{table}[t]
\centering
\caption{Quantitative comparisons across different fine-color adjusting strategies. ``W/o DPO'' indicates that no color adjustment training, and ``One-stage DPO'' refers to DPO training without the progressive strategy. Best results are \textbf{bolded.}}
\label{tab:abdpo}
\resizebox{\linewidth}{!}{%
\begin{tabular}{@{}lccccccc@{}}
\toprule
        
              &  DeQA & Q-Insight& VQ-R1 &CRI   & CRA   & CCS   & AES   \\ \midrule
\multicolumn{8}{c}{\textit{DIV2K-valid-augmented}}                     \\ \midrule
W/o DPO       & 3.932 & 3.752 & 4.344 & 55.80 & 71.85 & 79.44& 72.24 \\
SFT           &  4.236     &   3.939    &  4.656     &    70.95   &   78.65    &   84.02 & 79.80    \\
One-stage DPO &   \textbf{4.307}    &   3.972    &    4.692   &   77.05    &  79.35     &   \textbf{84.15} & 82.46    \\
Pro-DPO         &    4.303   &    \textbf{3.978}   &  \textbf{4.728}     &    \textbf{80.15}   &   \textbf{79.70}    &   82.66 & \textbf{82.80}    \\ \midrule
\multicolumn{8}{c}{\textit{RealOldPhotos}}                           \\ \midrule
W/o DPO       & 4.105 & 3.811 & 4.381& 67.80 & 79.60 & 82.16& 78.62 \\
SFT           &   4.145    &   3.838    &     4.456  &   73.10    &   81.10    &  83.14 & 80.54     \\
One-stage DPO &   \textbf{4.207}    &   3.883    &  4.565     &    77.80   &   \textbf{82.40}   &   83.38 & 82.00    \\
Pro-DPO         &   4.202    &   \textbf{3.894}    &    \textbf{4.600}   &   \textbf{80.40}    &   81.60    &   \textbf{83.72} & \textbf{82.28}    \\ \bottomrule
\end{tabular}%
}
\vspace{-3mm}
\end{table}

\begin{figure}
    \centering
    \includegraphics[width=\linewidth]{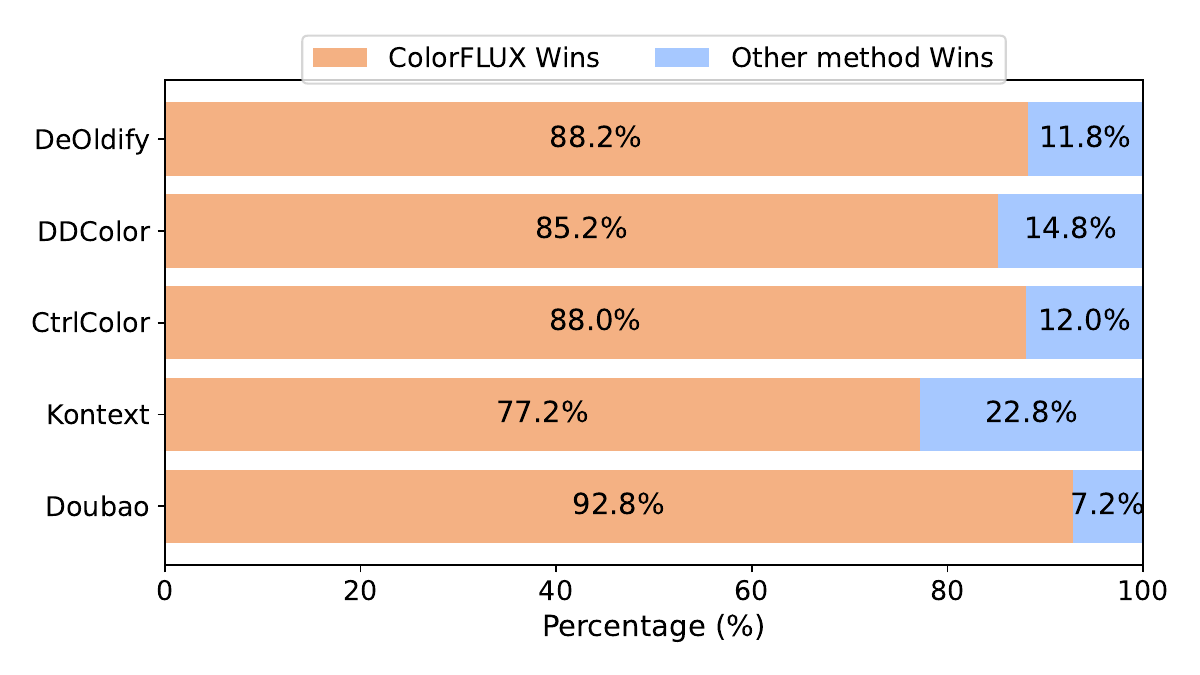}
    \caption{User study of ColorFLUX \emph{vs.} other methods on \textit{RealOldPhotos} among 10 participants.}
    \label{fig:winrate}
    \vspace{-4mm}
\end{figure}

\paragraph{Effectiveness of Progressive DPO Training} We compare our proposed progressive DPO strategy with three settings: (i) the model before DPO training (w/o DPO); (ii) supervised fune-tuning (SFT) using $(c, x^w)$ pairs; and (iii) one-stage DPO, where a wide range of color augmentations is applied throughout training. As demonstrated in Tab.~\ref{tab:abdpo}, our proposed Pro-DPO achieves almost the best results across all metrics. Compared to one-stage DPO, our progressive strategy enables the model to learn more fine-grained color adjustments, resulting in colorization outputs that better align with modern color preferences. Additionally, as illustrated in Fig.~\ref{fig:abdpo}, Pro-DPO produces the most vivid and natural colors, particularly in fine-grained regions such as individual fruits. This highlights the advantage of our progressive training strategy in capturing subtle color preferences and enhancing visual realism.

\subsection{User Study}

We invited 10 experienced researchers in the image processing field to participate in our user study. Each participant was asked to conduct pairwise comparisons between the results of our method and those of competing methods, and to select the more preferred one. The comparisons were conducted on all images from \textit{RealOldPhotos}. To avoid bias, we randomly shuffled the presentation order of the pairs for each participant. The results are presented in Fig.~\ref{fig:winrate}. As observed, our method achieves a significantly higher win rate over all baselines, demonstrating its strong advantage in generating visually preferred colorization results. More details are provided in Sec.~\ref{sec:userstudy}.

\section{Conclusion}

In this paper, we present a novel old photo colorization framework, dubbed ColorFLUX, that leverages a structure-color decoupling strategy and a progressive Direct Preference Optimization approach to address the unique challenges of colorizing degraded historical images. By incorporating visual semantic prompts, our method bypasses the limitations of text-based guidance and captures fine-grained semantics directly from the input. Extensive experiments on both synthetic and real datasets demonstrate that our approach achieves superior structural consistency and color vividness, surpassing existing state-of-the-art methods. However, our current implementation relies on the FLUX model, which limits its deployment on resource-constrained devices. In future work, we plan to explore lightweight alternatives to enable broader applicability while maintaining colorization performance.

{
    \small
    \bibliographystyle{ieeenat_fullname}
    \bibliography{main}
}

\clearpage
\maketitlesupplementary

\section{Training Details about ColorFLUX}
\label{sec:details}

\begin{figure*}[!ht]
    \centering
    \includegraphics[width=1\linewidth]{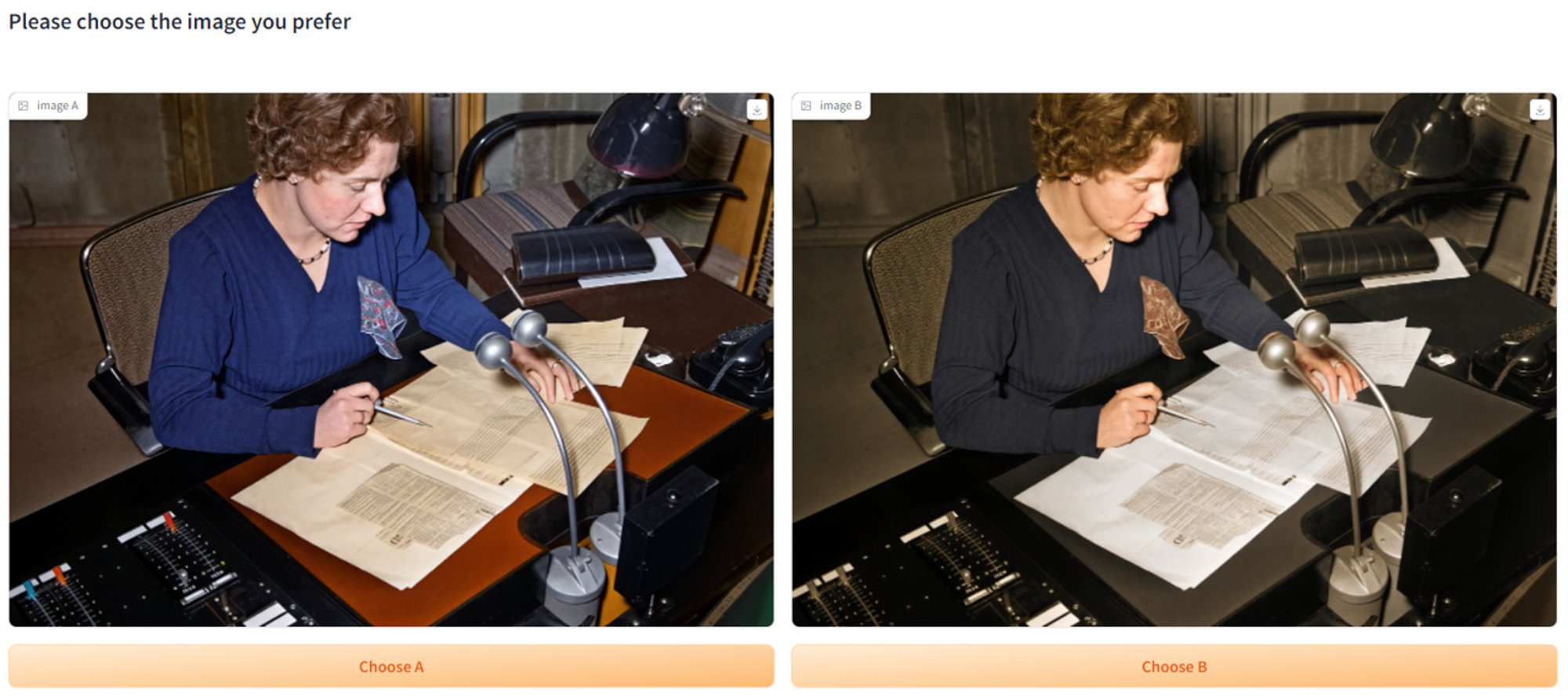}
    \caption{A screenshot of our user study.}
    \label{fig:userstudy}
\end{figure*}

\begin{figure}[!ht]
    \centering
    \includegraphics[width=1\linewidth]{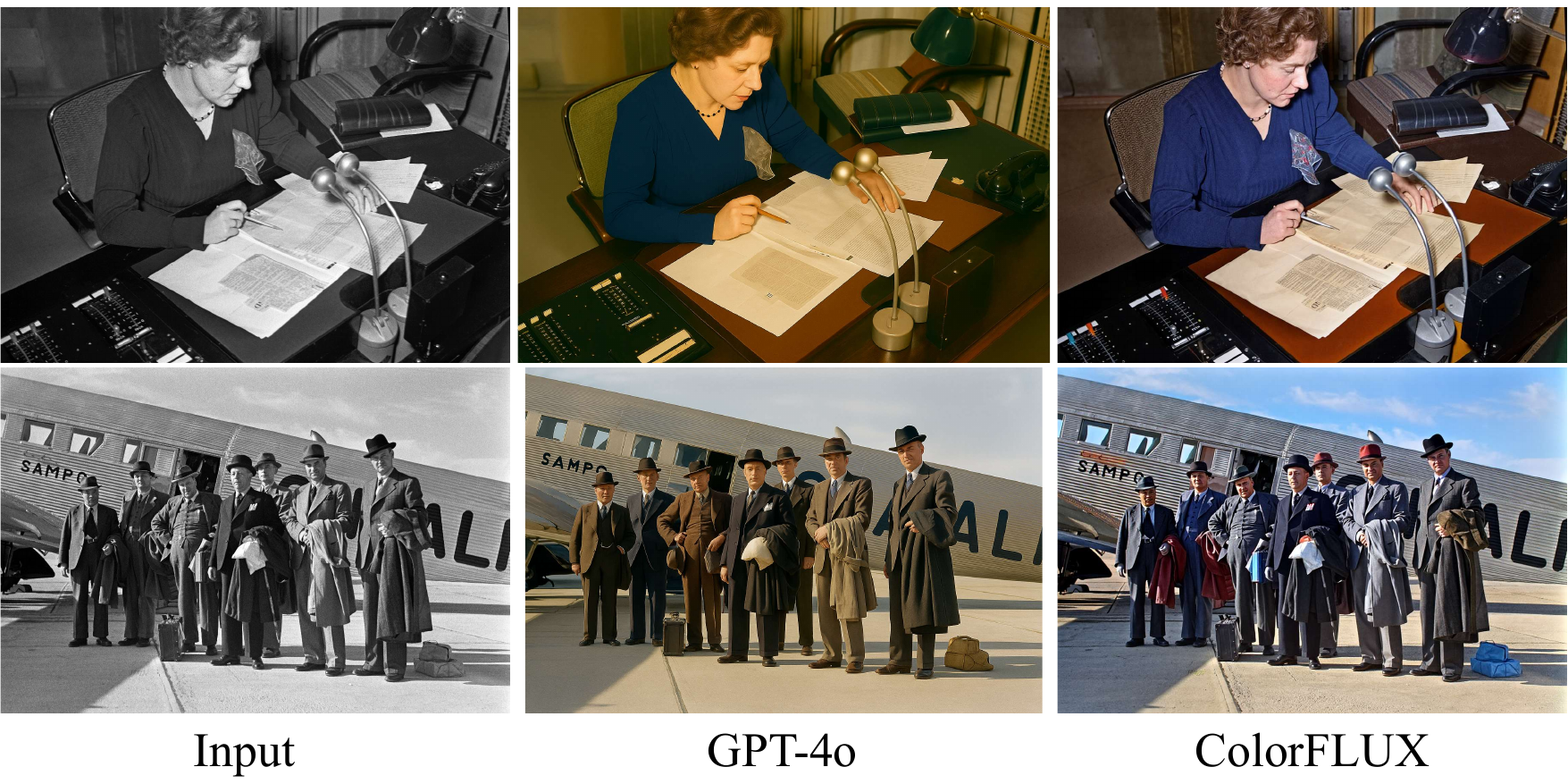}
    \vspace{-10mm}
    \caption{Visual comparisons between ColorFLUX and closed-source GPT-4o. Zoom in for better views. GPT results exhibit noticeable structural inconsistencies, particularly in regions such as fingers and faces.}
    \label{fig:suppgpt}
\end{figure}

\subsection{Structure Consistency Training}

We initiate ControlNet by replicate 4 double-stream and 10 single-stream MM-DiT blocks of the pre-trained FLUX~\cite{flux2024}, following the default FLUX ControlNet settings. For dataset, we utilize a combination of LSDIR~\cite{li2023lsdir} and the first 10k images from FFHQ~\cite{karras2019styleFFHQ}, following the common approaches in recent image restoration works~\cite{wu2024osediff,chen2025adversarialadcsr}. We train ControlNet with a batch size of 64 and a learning rate of 1e-5 for 5 epochs. 

\subsection{Basic Color Learning} Since LSDIR was originally proposed for image restoration, it includes some grayscale or very limited colored images, which can negatively affect our colorization task. To filter out these images, we employ the statistic-based colorfulness metric proposed by Hasler and Süsstrunk~\cite{hasler2003measuringcolorfulness}. By setting the threshold to 15, we ultimately retain 76956 (out of 84991) images with sufficient color diversity. We \textit{fix ControlNet} and only optimize the Redux model for 2 epochs with a batch size of 64 and a learning rate of 1e-5. We set the weighting parameter $\alpha=0.1$.

\subsection{Fine-Color Adjustment} 
Since DPO training is more sensitive to color quality, we instead use the higher-quality datasets DIV2K~\cite{DIV2K} and Flickr2K~\cite{Flickr2K}. Additionally, beyond the colorfulness-based filtering metric, we further convert each image to the HSV color space and compute the mean values of the $S$ and $V$ channels to represent saturation and brightness, respectively. We then set thresholds of $0.3\sim 0.7$ for saturation and $0.4\sim 0.8$ for brightness to ensure the selected images exhibit sufficient chromatic diversity and proper exposure. This results in 1558 remained images. For color augmentation, we focus on Colorfulness, Brightness, and Contrast since these factors are the most related to the fading effects in old photos. To realize these augmentations, we utilize \texttt{Pillow.ImageEnhance} python package. In the first stage, we optimize LoRAs with a rank of 32 using a batch size of 32 and a learning rate of 4e-5 for 2 epochs. We set the augmentation range as [0.5, 0.8] for this stage. The same procedure is applied in the second stage, except that we adopt milder color augmentation range [0.75, 0.95] and a lower learning rate 1e-5. We adopt a fix $\beta=1000$ during the two stage training.

\section{More Details about Qwen-score}
\label{sec:qwen}

How to assess the quality of an image, \ie, IQA, has long been a fundamental problem in low-level vision research. Despite numerous metrics~\cite{you2025teachingdeqa,li2025qinsight,yang2022maniqa,wu2025visualquality}, there is still no single metric that can comprehensively evaluate the performance of image colorization. Inspired by GPT-score~\cite{tong2025gptscore}, which leverages the rich prior knowledge of large vision-language foundation models to assess the quality of generated outputs, we propose to utilize the prior knowledge embedded in Qwen2.5-VL-72B~\cite{Qwen2.5-VL} to evaluate the quality of image colorization results. Additionally, Qwen has been adopted as the base model in several recent IQA works~\cite{wu2025visualquality,li2025qinsight}, further demonstrating its potential in IQA and supporting the rationale behind our choice. We focus on six aspects of colorized images:
\begin{itemize}
\item \textbf{Color RIchness (CRI)}: Assesses the diversity and vividness of colors in the image.
\item \textbf{Color RAtionality (CRA)}: Evaluates whether the colors are harmonious and consistent with real-world semantics.
\item \textbf{Color ConSistency (CCS)}: Measures whether the image suffers from color bleeding across regions with similar semantics.
\item \textbf{Structural ConSistency (SCS)}: Evaluates the structural fidelity between the input grayscale image and the colorized result.
\item \textbf{AESthetics (AES)}: Reflects the overall visual appeal of the colorized image.
\item \textbf{OverAll (OA)}: Provides a holistic score by considering all of the above aspects.
\end{itemize}

\begin{figure*}[t]
    \centering
    \includegraphics[width=0.8\linewidth]{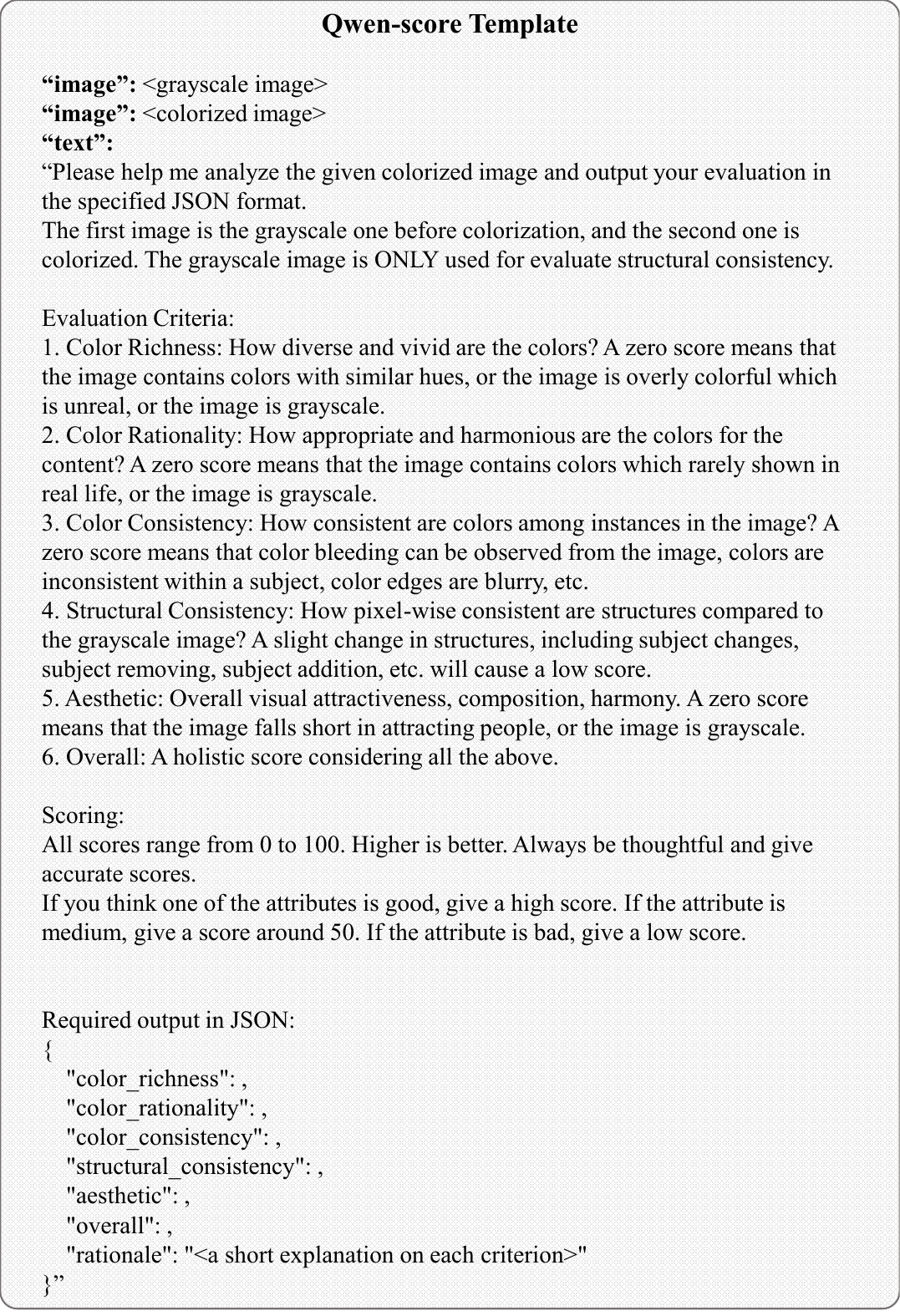}
    \caption{The template of our proposed Qwen-score.}
    \label{fig:qwentemp}
\end{figure*}

We provide the prompt template in Fig.~\ref{fig:qwentemp}.

\section{User Study}
\label{sec:userstudy}

To ensure a more thorough and reliable comparison with existing methods, we conducted a user study involving 10 experts specializing in low-level vision. Each expert was asked to sequentially compare image pairs from the \textit{RealOldPhotos} dataset and select the one they preferred in each pair. For each comparison, one image was always the result generated by ColorFLUX, while the other was randomly selected from one of the five competing methods (as listed in Tab.~\ref{tab:maintable}). To prevent potential bias, the left–right order of images in each pair was randomized, and the presentation order of all 250 pairs was also shuffled to avoid repeated appearances of the same image in close succession. After all 10 experts completed the evaluation of 250 pairs, we computed the win rate for ColorFLUX based on the total number of times it was preferred over each baseline method. We demonstrate a screenshot of a pair comparison in Fig.~\ref{fig:userstudy}.

\section{More Visual Results}

\begin{figure*}[t]
    \centering
    \includegraphics[width=0.7\linewidth]{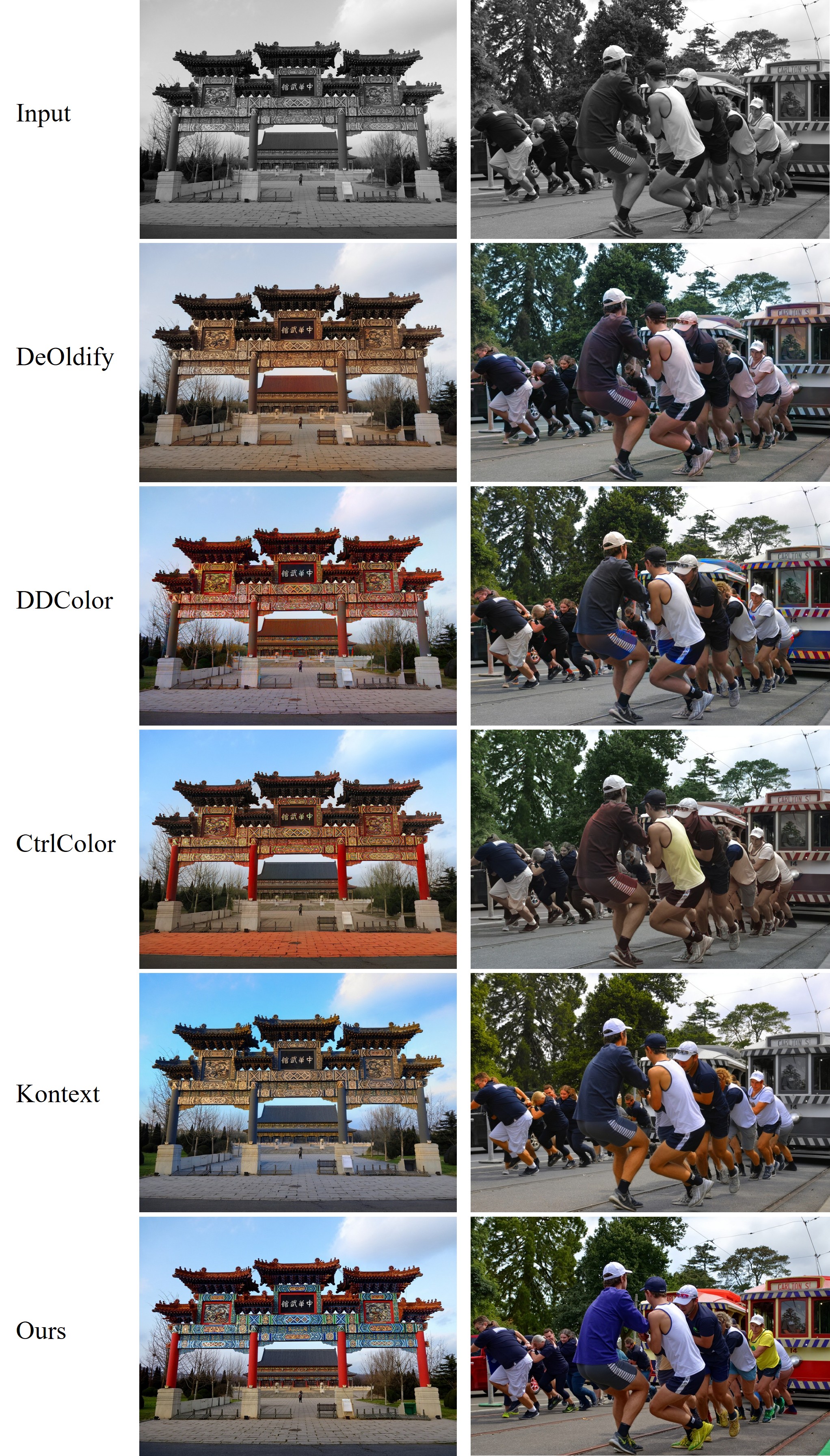}
    \caption{Visual comparisons between ColorFLUX and other methods on \textit{DIV2K-valid-synthesized} dataset.}
    \label{fig:suppdiv2k1}
\end{figure*}

\begin{figure*}[t]
    \centering
    \includegraphics[width=0.75\linewidth]{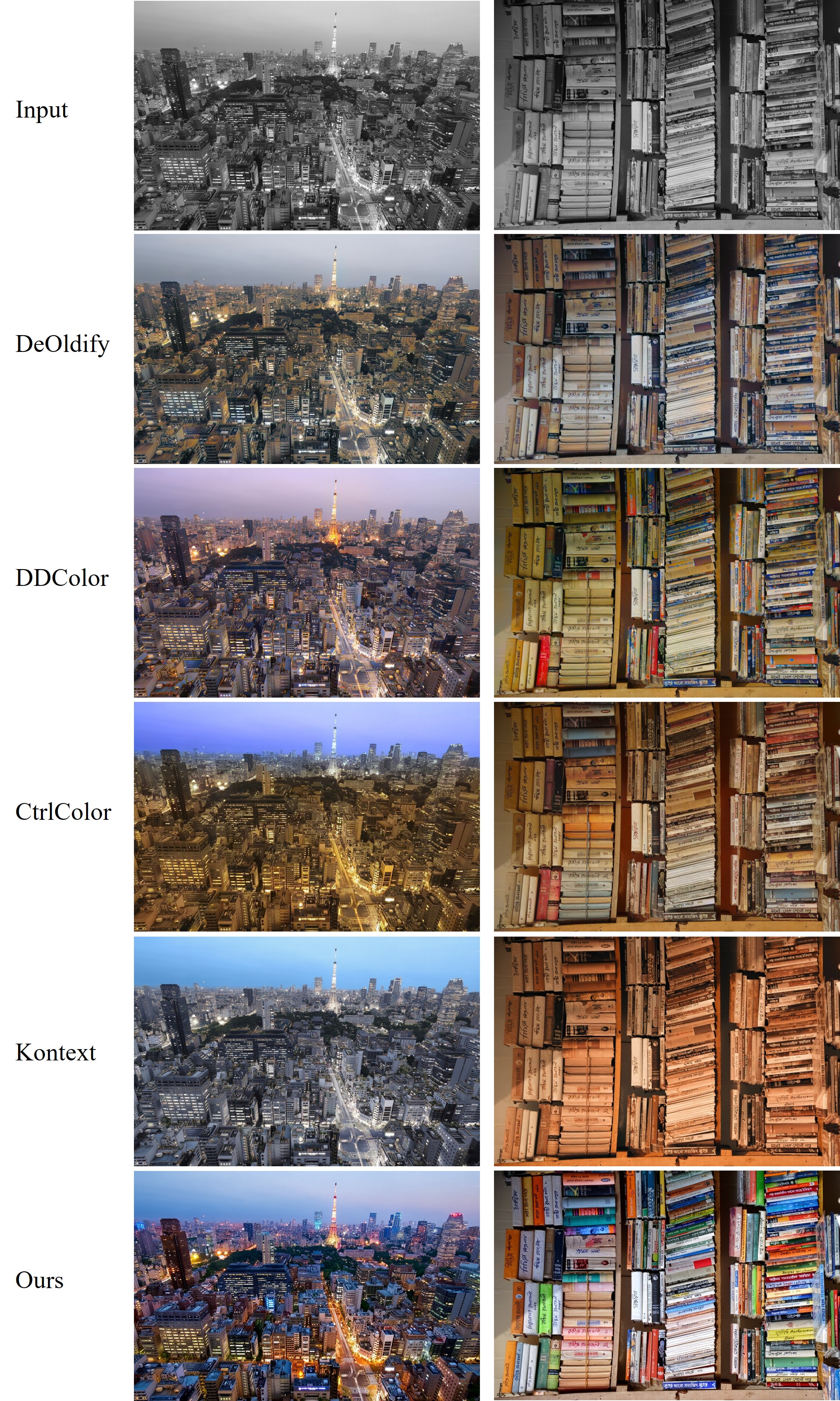}
    \caption{Visual comparisons between ColorFLUX and other methods on \textit{DIV2K-valid-synthesized} dataset.}
    \label{fig:suppdiv2k2}
\end{figure*}

\begin{figure*}[t]
    \centering
    \includegraphics[width=0.85\linewidth]{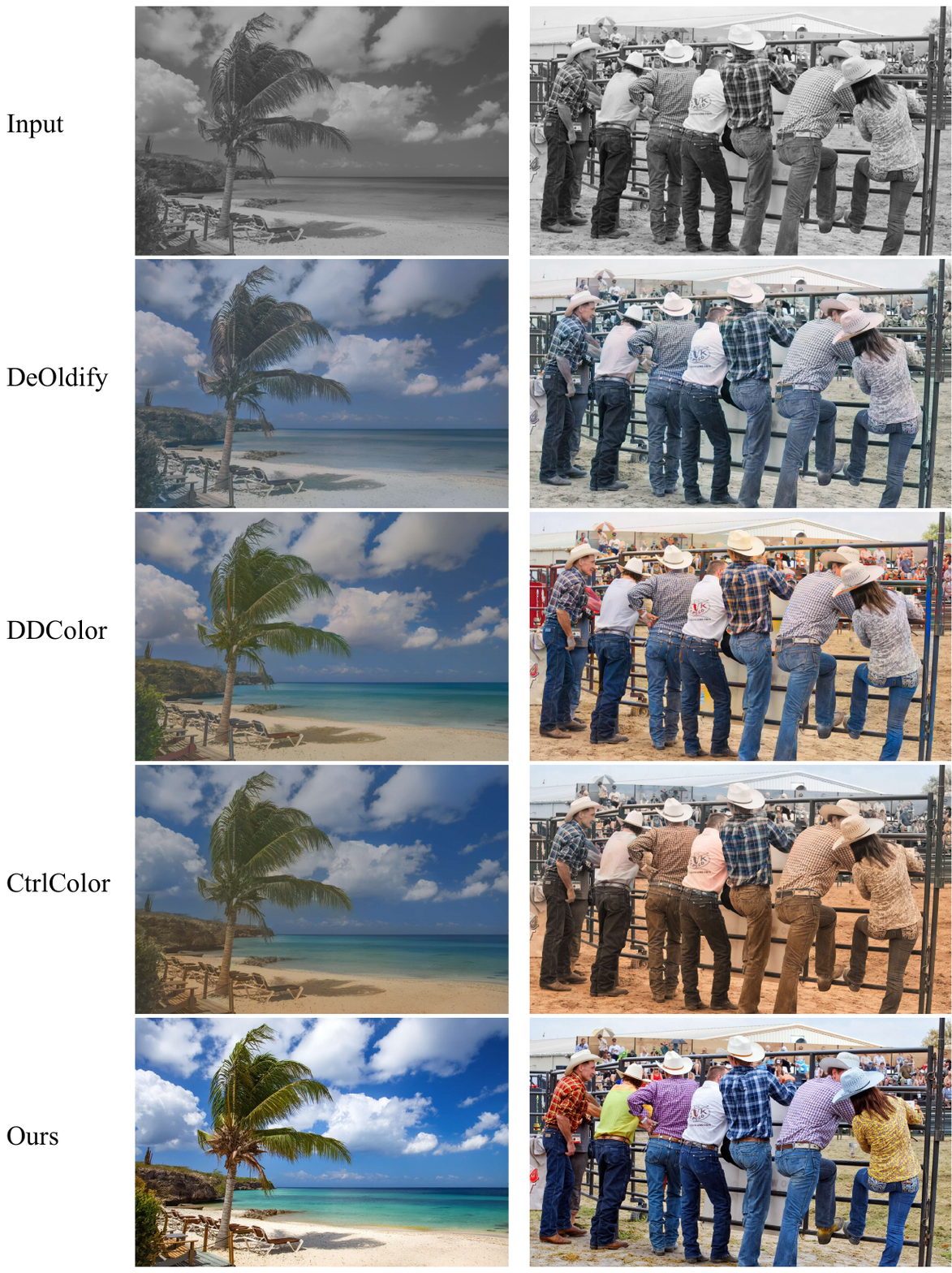}
    \caption{Visual comparisons between ColorFLUX and other methods on \textit{DIV2K-valid-augmented} dataset.}
    \label{fig:suppdiv2k3}
\end{figure*}

\begin{figure*}[t]
    \centering
    \includegraphics[width=0.9\linewidth]{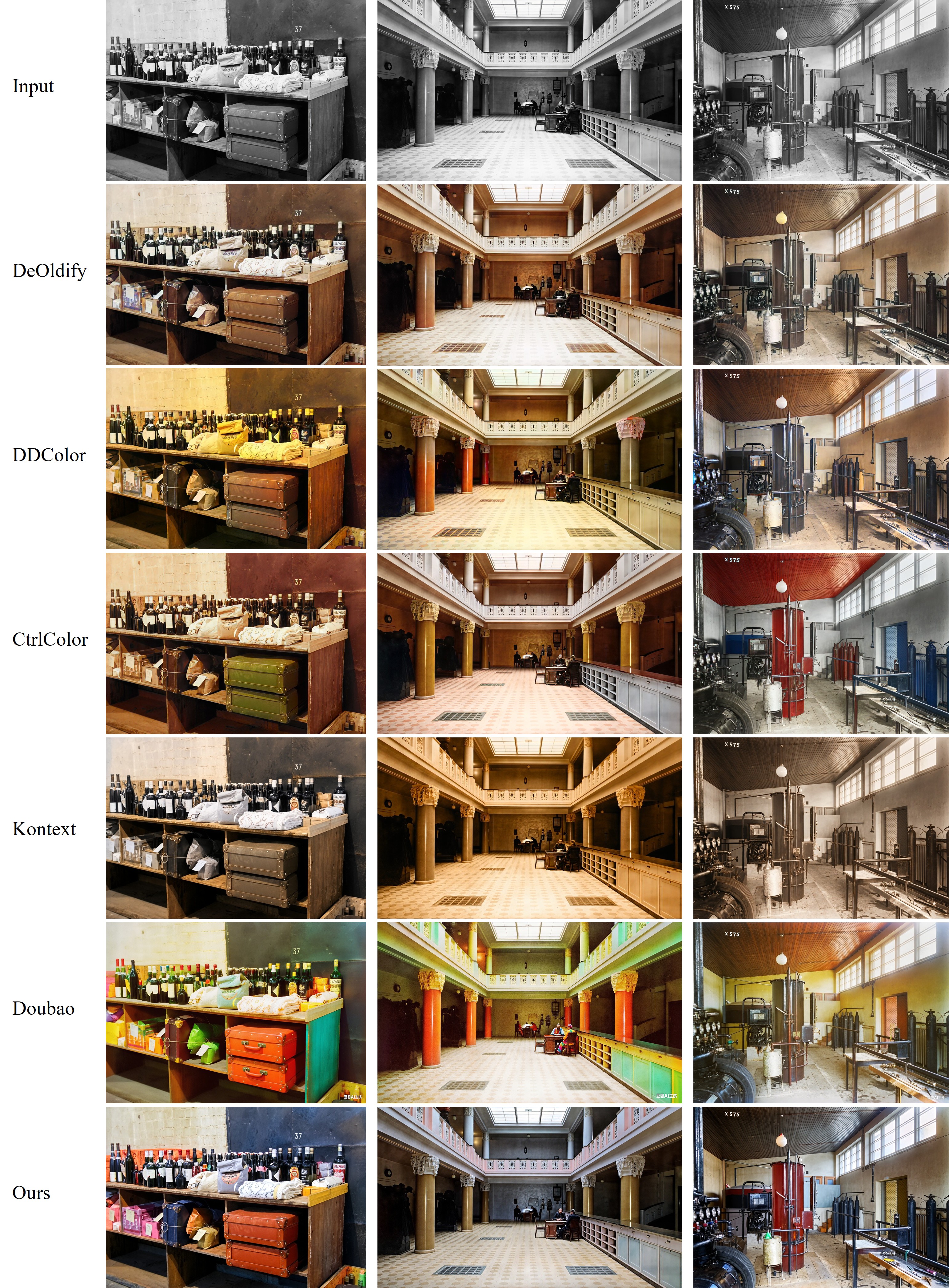}
    \caption{Visual comparisons between ColorFLUX and other methods on \textit{RealOldPhotos} dataset.}
    \label{fig:supprealold1}
\end{figure*}

\begin{figure*}[t]
    \centering
    \includegraphics[width=0.85\linewidth]{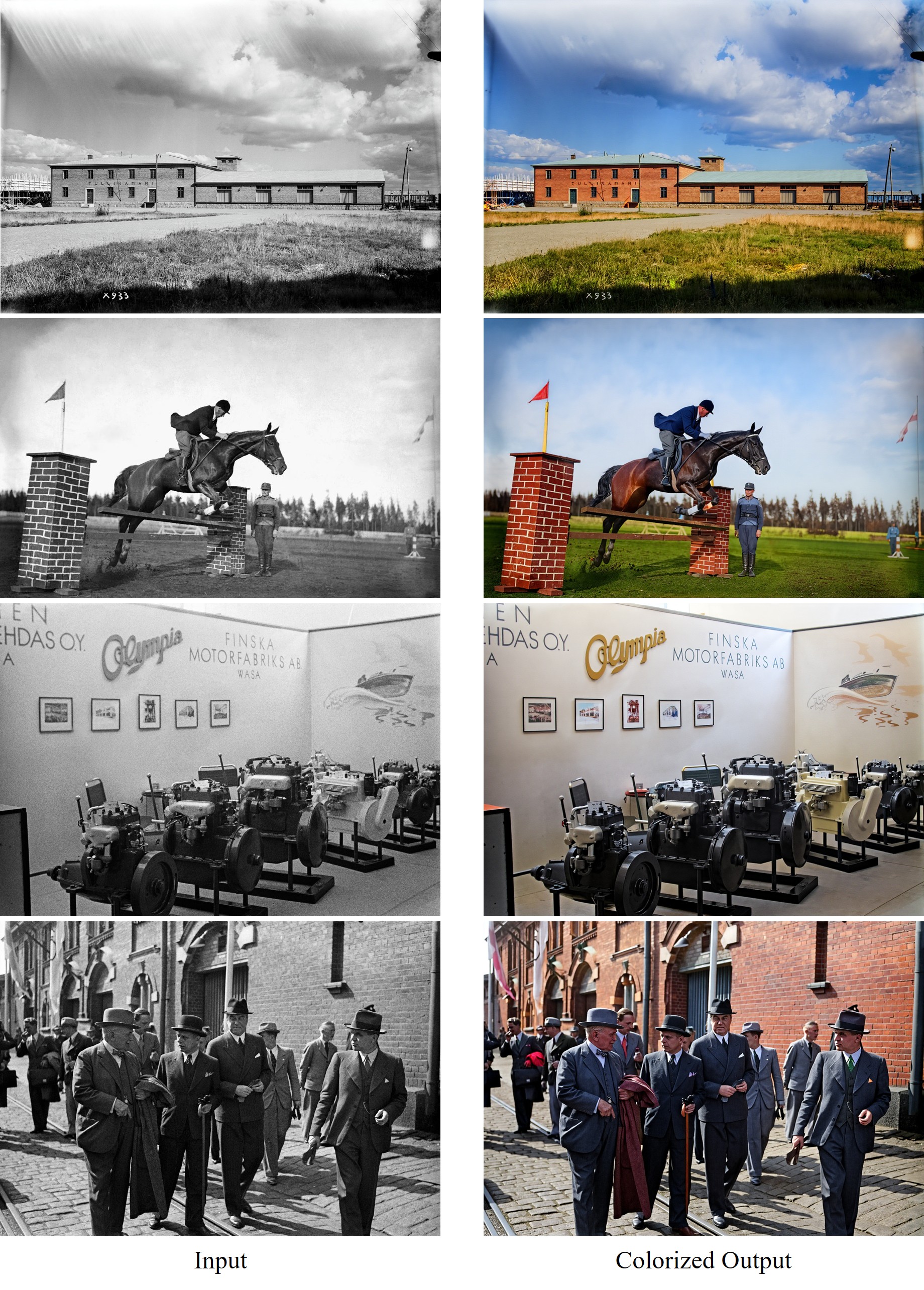}
    \caption{Colorization results of ColorFLUX on \textit{RealOldPhotos}.}
    \label{fig:suppour2}
\end{figure*}

\begin{figure*}[t]
    \centering
    \includegraphics[width=0.85\linewidth]{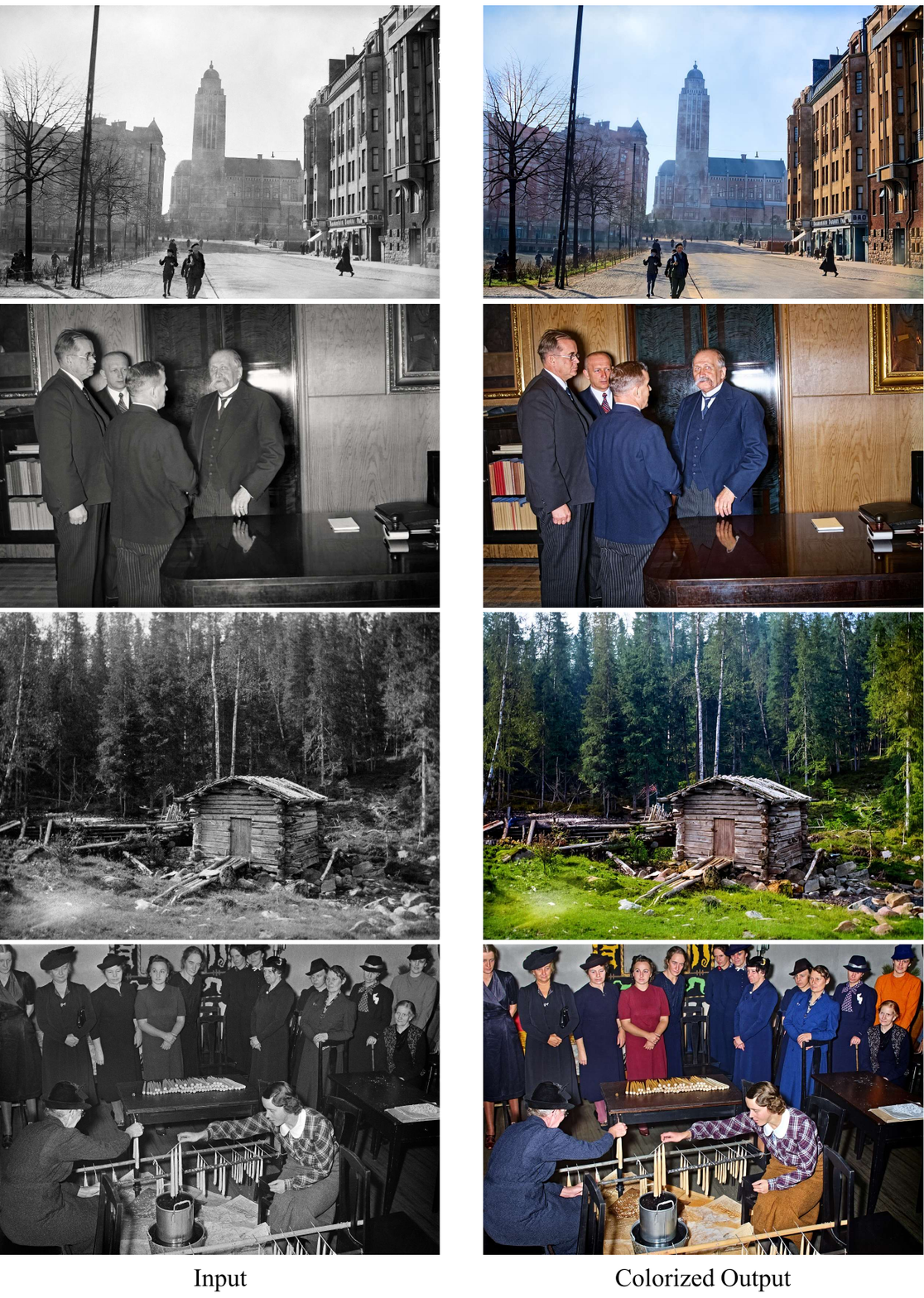}
    \caption{Colorization results of ColorFLUX on \textit{RealOldPhotos}.}
    \label{fig:suppour3}
\end{figure*}

\begin{figure*}[t]
    \centering
    \includegraphics[width=0.85\linewidth]{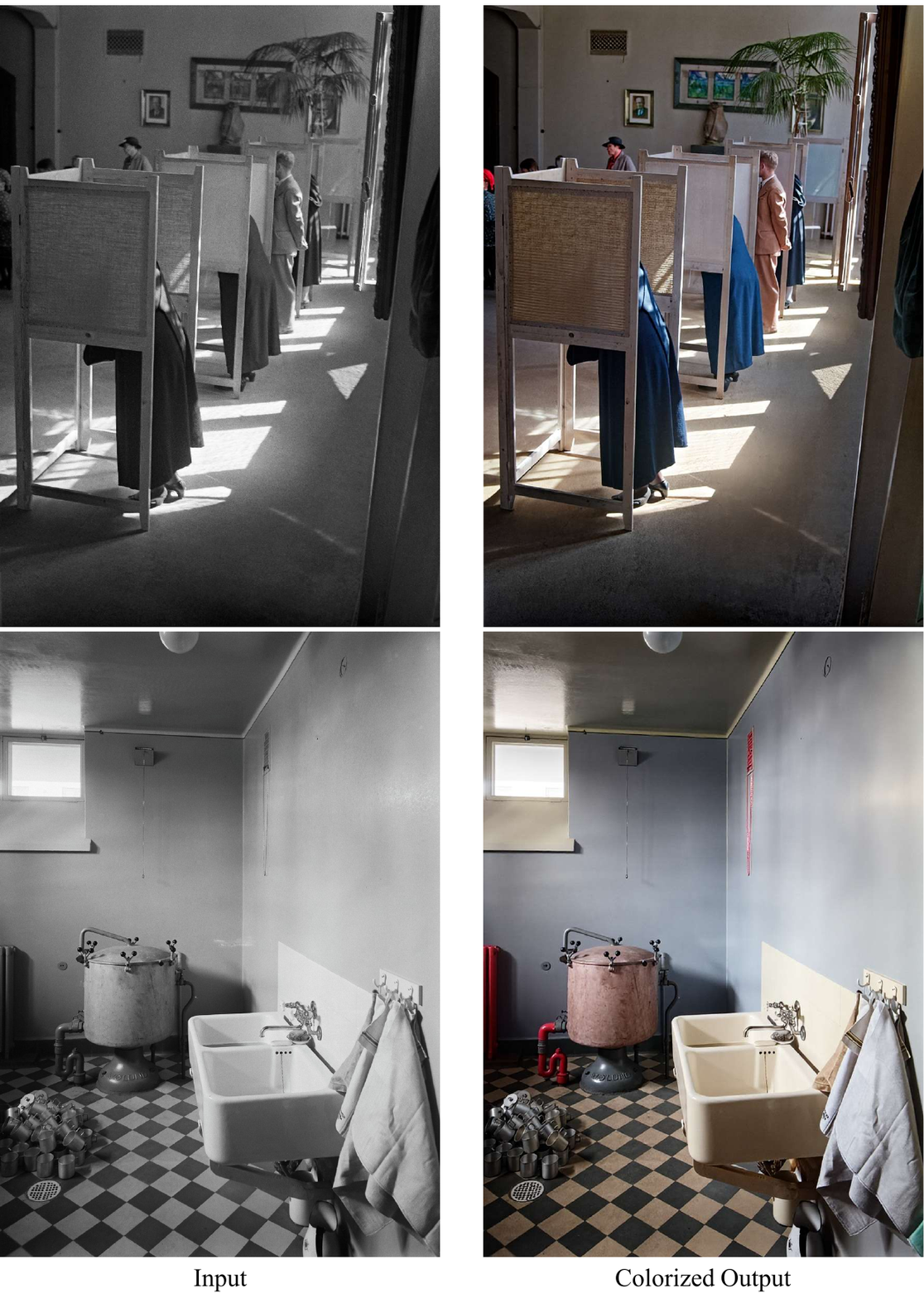}
    \caption{Colorization results of ColorFLUX on \textit{RealOldPhotos}.}
    \label{fig:suppour4}
\end{figure*}

In this section, we demonstrate more visual comparisons and high-quality colorization results by our proposed ColorFLUX. We depict the comparisons between our method with the powerful closed-source model GPT-4o in Fig.~\ref{fig:suppgpt}. However, the colorization results generated by GPT exhibit noticeable structural inconsistencies, particularly in regions such as fingers and faces. This observation suggests that our method even outperforms commercial-grade models. We provide visual comparisons on \textit{DIV2K-valid-synthesized} in Fig.~\ref{fig:suppdiv2k1} and Fig.~\ref{fig:suppdiv2k2} to evaluate the colorization abilities of ColorFLUX against other methods under traditional image colorization settings (\ie, \texttt{RGB2GRAY}). As demonstrated, ColorFLUX achieves the most realistic and vivid colorization results across all grayscale pictures. As illustrated in Fig.~\ref{fig:suppdiv2k3}, ColorFLUX demonstrates the strongest robustness against synthesized fading effects (\eg, saturation reduction), whereas other methods fail to generate images with natural and realistic color tones. Moreover, in the rightmost column, ColorFLUX is the only method that avoids color bleeding on the leftmost person’s arm, an issue clearly observed in all other approaches. These results highlight the effectiveness of our visual semantic prompt design and the proposed progressive DPO training strategy. Additionally, we provide more high-resolution, high-quality colorization results in Fig.~\ref{fig:supprealold1}, Fig.~\ref{fig:suppour2}, Fig.~\ref{fig:suppour3}, and Fig.~\ref{fig:suppour4}. These results demonstrate that our method achieves semantically consistent and visually realistic colorization with rich hues, regardless of scene complexity.

\clearpage
\onecolumn

\end{document}